\documentclass[%
 reprint,
 amsmath,amssymb,
 aps,
pre,
superscriptaddress,
]{revtex4-2}

\usepackage[utf8]{inputenc}
\usepackage[T1]{fontenc}
\usepackage{booktabs}
\usepackage{amsmath}
\usepackage{blkarray, bigstrut}
\usepackage{xparse}
\usepackage{multirow}
\usepackage{comment}
\usepackage{xfrac}

\usepackage{subcaption}
\usepackage{graphicx}% Include Fig.files
\usepackage{dcolumn}% Align table columns on decimal point
\usepackage{bm}% bold math
\usepackage{hyperref}% add hypertext capabilities
\usepackage{soul}
\hypersetup{colorlinks=false, linktoc=all, linkcolor=blue, linktocpage, citecolor=blue, hidelinks}
\makeatletter
\renewcommand{\fnum@figure}{Fig. \thefigure}
\makeatother

\usepackage{xfrac}  
\usepackage[normalem]{ulem}
\usepackage{lineno}
\usepackage[usenames,dvipsnames]{xcolor}
\modulolinenumbers[5]
\begin{document}

%\preprint{APS/123-QED}

%alternativas de nomes
%\title{Introducing Dynamics into the Prisoner's Dilemma Game with Reinforcement Learning}
%\title{A Study of Dilution and Diffusion in the Spatial Prisoner's Dilemma with Reinforcement Learning}
\title{Dilution, Diffusion and Symbiosis in the Spatial Prisoner’s Dilemma with Reinforcement Learning}
%ou talvez esse
%\title{Agents Can Learn to Cooperate When Given Enough Space In the Spatial Prisoner's Dilemma with Q-Learning}

% \altaffiliation[Also at ]{Instituto de F\'{i}sica, Universidade Federal do Rio Grande do Sul, Brazil.}%Lines break automatically or can be forced with \\

\author{Gustavo C. Mangold}%
 %\email{heitor.fernandes@ufrgs.br}

\author{Heitor C. M. Fernandes}%
 %\email{heitor.fernandes@ufrgs.br}

\author{Mendeli H. Vainstein}%
 %\email{vainstein@if.ufrgs.br}

\affiliation{%
 Instituto de Física, Universidade Federal do Rio Grande do Sul, Porto Alegre, RS, Brazil
}%

%\author{Mateus Guimarães}%
 %\email{mateus.guimaraes048@gmail.com}

 %\affiliation{%
 %Instituto de Ciências Básicas da Saúde, Universidade Federal do Rio Grande do Sul, Porto Alegre, RS, Brazil
%}%
%\collaboration{}%\noaffiliation

\date{\today}% It is always \today, today,

\begin{abstract}
Recent studies in the spatial prisoner's dilemma games with reinforcement learning have shown that static agents can learn to cooperate through a diverse sort of mechanisms, including noise injection, different types of learning algorithms and neighbours' payoff knowledge.
  In this work, using an independent multi-agent Q-learning algorithm, we study the effects of dilution and mobility in the spatial version of the prisoner's dilemma.
  Within this setting, different possible actions for the algorithm are defined, connecting with previous results on the classical, non-reinforcement learning spatial prisoner's dilemma, showcasing the versatility of the algorithm in modeling different game-theoretical scenarios and the benchmarking potential of this approach.
  As a result, a range of effects is observed, including evidence that games with fixed update rules can be qualitatively equivalent to those with learned ones, as well as the emergence of a symbiotic mutualistic effect between populations that forms when multiple actions are defined.
\end{abstract}

\maketitle

\section{\label{sec:intro}Introduction}
In nature, although individuals might gain more short term rewards if they are selfish, cooperation emerges as an alternative that can promote the maintenance of a species through time, such as in honey bee populations which coordinate to form a super organism that can resist and thrive through diversity and cooperation \cite{seeley2011honeybee, michener1974social, frisch1993dance}.
Cooperative behaviour can also appear in interactions between different species that work together, as in symbiotic relationships that emerge, for example, in bacteria that live within insects~\cite{hosokawa2016obligate}, and in plant-animal interactions \cite{bronstein2006evolution, vazquez2009uniting, vaudo2024pollen}.

If we want to understand how cooperation appears in scenarios where selfishness is usually more viable, we need a model that encapsulates the behaviour associated with interactions between individuals. 
One way to model simple scenarios that do this while also mimicking natural behaviour is through the Prisoner's Dilemma (PD) rule-set.  Its origin is 
%The Prisoner's Dilemma (PD) game  arises from 
 a classical anecdote in which two individuals (usually called players), faced with a trial, are presented the options to either cooperate with one another or to  defect~\cite{rapport1965prisoner}, with each pair of actions having associated compensations.
In its simplest form, the two players participating in the game get a reward ($R$) if they mutually cooperate and a punishment ($P$) if the common choice is defection.
If the choices differ, the  player choosing to cooperate receives a \textit{sucker's payoff} ($S$), while the defector receives the temptation payoff ($T$).
The PD is characterized by the inequalities $T > R > P > S$ and $2R > T+S$~\cite{nowak1992evolutionary, axelrod1981evolution}.
In its spatial version, where evolutionary aspects more  closely resemble biological settings \cite{vincent2005evolutionary}, interactions occur between neighbours in a certain topology, equipped with a reward system derived from the usual PD payoffs. 
It has been shown that these evolutionary dynamics games in spatial settings present different cooperation levels among agents depending on the choice of topological structure \cite{FLORES2022112744}, on strategy update rules \cite{takesue2019effects, kaiping2014nonequivalence}, spatial disorder \cite{vainstein2001disordered}, asymmetry and heterogeneity \cite{pacheco2009population, mcavoy2015structural, du2024asymmetric, flores2023heterogeneous}, and mobility \cite{vainstein2007does}.

Considering the recent advances in reinforcement learning -- an inherently self-interested algorithm that encourages agents to learn which actions yield the highest rewards~\cite{sutton2018reinforcement} -- we may question the suitability of such algorithms for modeling agents engaged in games that can favor selfishness, such as the PD.
%\st{Considering the recent advances in reinforcement learning,  is based on an inherently selfish algorithm that induces players to learn which actions return the greatest rewards}~\cite{sutton2018reinforcement}\st{, we can ask ourselves about the suitability of such algorithms to model agents that play a game with a specific rule set.} 
We can further inquire if cooperation can persist when players in the PD learn strategies with reinforcement learning, and which mechanisms, such as clustering, coordination or symbiosis, can help  maintain cooperation. 
Reinforcement learning, in these kinds of settings, was originally used for the iterated version of the PD game, where players learned to play optimally against different strategies \cite{SANDHOLM1996147}, such as tit-for-tat~\cite{axelrod1981emergence}. Previous studies have focused on the impact of learning agents in spatial games, where each player updates their strategy using an independent multi-agent Q-learning algorithm~\cite{watkins1992q}. This approach has been applied to spatial versions of the PD~\cite{zhang2019collective, zhang2020oscillatory, wang2022levy}, including scenarios involving punishment~\cite{zhao2024emergence, yan2024periodic}, as well as to the Public Goods Game~\cite{wang2023synergistic, zhang2024exploring}.
 %an \textcolor{red}{update rule OU strategy} for each player is learned via an independent multi-agent Q-learning algorithm \cite{watkins1992q}, with applications in the spatial prisoner's dilemma game \cite{zhang2019collective, zhang2020oscillatory, wang2022levy}, in some cases with punishment \cite{zhao2024emergence, yan2024periodic}, and in the public goods game \cite{wang2023synergistic, zhang2024exploring}.
%Reinforcement learning, in these kinds of settings, was originally used for the iterated version of the prisoner's dilemma game, where players learned to play optimally against different strategies \cite{SANDHOLM1996147}, such as tit-for-tat \cite{axelrod1981emergence}.
When this type of algorithm is implemented on a lattice with nearest-neighbor interactions in the context of the PD, the result is typically the dominance of defectors and very low levels of cooperation as noted in~\cite{wang2022levy}. This outcome is expected in such settings, where players have little to no information about their surroundings -- cooperators lack topological awareness and are therefore unable to form clusters, making them vulnerable to invasion by defectors.
%\textcolor{red}{On the spatial side, if we run this type of algorithm on a lattice with nearest neighbours interactions in the prisoner's dilemma context, as pointed by \cite{wang2022levy}, an emergence of defectors, with very low levels of cooperation, is observed. -- MELHORAR e/ou juntar com acima. Colocar antes? -- This is expected in such a setting, where players have little to no knowledge of their surroundings, as cooperators have no topological awareness and thus can't cluster together to resist invasion from defectors.}  %revisar isso, talvez achar alguma citaçao

With this in mind, we might ask how defects in the lattice and player mobility can influence the outcome of the learning process. Previous studies have addressed this question without reinforcement learning, using predefined update rules, and found that, in most cases, a certain level of mobility promotes cooperation~\cite{vainstein2007does, meloni2009effects}.  %\textcolor{red}{\st{In light of this, we consider spatial games on a square lattice occupied by a fixed density of players, $\rho$, leaving a fraction $1 -\rho$ of empty spaces. We will investigate the effects of this dilution, as well as the impact of diffusive mobility.} -- MODELO}
%With all of this in mind, we might ask ourselves how defects in the network and mobility can alter the outcome of the learning process, as previous works have already answered this same question without reinforcement learning, with predefined update rules and found that, in most cases, a certain level of mobility does increase cooperation \cite{vainstein2007does, meloni2009effects}. %revisar isso aqui mendeli
%We thus consider spatial games in a square lattice occupied with an exact density of players $\rho$, leaving thus a fraction $1-\rho$ of empty spaces behind, where we will study the effects of this dilution and also of diffusive mobility.

In our study, we use a multi-agent reinforcement learning (MARL) algorithm -- specifically, an independent multi-agent Q-learning approach \cite{tan1993multi, busoniu2008comprehensive} implemented in an online fashion \cite{wei2017online}. In this setup, each agent updates its state after receiving a reward. The algorithm exhibits single-agent characteristics when viewed from the perspective of an individual agent, while displaying multi-agent dynamics when considering the environment as a whole.
%For our studies in this environment, we employ a multi-agent reinforcement learning algorithm (MARL), or, more specifically, an independent multi-agent Q-learning algorithm \cite{tan1993multi, busoniu2008comprehensive} in an online fashion \cite{wei2017online}, which has its state updated after a reward is received and is composed of single-agent aspects when looking at only one agent, and multi-agent behaviour when looking at the whole environment.
This algorithm has recently been referred to as \textit{self-regarding} in the evolutionary dynamics with reinforcement learning literature  \cite{wang2022levy, wang2023synergistic, (fermi-update-q-learning)guo2022effect}. However, for the sake of terminological precision, we will use the more established term and refer to the algorithm throughout the text as independent multi-agent Q-learning.
%This algorithm was re-coined as \textit{self-regarding} in the recent  evolutionary dynamics with reinforcement learning literature \cite{wang2022levy, wang2023synergistic, (fermi-update-q-learning)guo2022effect}, but, for nomenclature exactitude, we will use the most established one and mention the algorithm throughout the text as independent multi-agent Q-Learning.
It is also worth noting that if the agents were designed to maximize a global reward, the simulation could be framed as a cooperative Markov game~\cite{matignon2012independent}. However, this is not the case in our simulations, where agents act selfishly, aiming solely to maximize their individual rewards.
%We must also note that if the agents' goal is set to increase the global reward, the simulation could compose a cooperative Markov game \cite{matignon2012independent}.  This is not exactly the case in our simulations, as agents work selfishly towards increasing only their own reward. 

Having defined the algorithm to be used, we  define which states and actions will be available to the agents.
For this, we begin by fixing a set of states $\mathbf{s}$, which will determine only whether the player is a cooperator or defector, and we perform case studies by varying the action space $\mathbf{a}$. 
This allows us to study how agents behave and cooperation is sustained with a series of different sets of actions in the same type of environment. We begin with actions that consider only strategy changes, i.e., with static players located on a diluted lattice~\cite{vainstein2001disordered} and that have no spatial or neighbourhood awareness. 
%\textcolor{orange}{\st{can only choose to change their strategy or not.}} 
%\textcolor{orange}{RESULTADO?
%It is observed, then, that players in a static agents attain a mixed state of cooperators and defectors in every density, but a screening effect appears when the density decreases and the environment becomes more diluted, due to isolation of players and asymmetry in the interactions, increasing cooperation levels.}
In this setting, we then introduce diffusive mobility with a mobility rate as a simple addition to the first action set.
%\textcolor{orange}{RESULTADO? This new action shows no real change in the behaviour of agents, explained by the lack of knowledge of players.}

%\textcolor{orange}{RESULTADO? Seeing the little effects of mobility with the previous set of actions, } 
After studying this scenario, the \textit{no-knowledge} case, we create new actions based on the literature of the PD game, which will give the players  awareness of their surroundings.
The first novel action is to \textit{copy-the-best}  strategy in the neighbourhood \cite{nowak1992evolutionary}. 
%Together with the action to move, this action forms a set that, when agents are presented with it, results in cooperation increasing for low mobility and densities around the percolation limit of the network, and in total defection for mobility close to one, which connects with results from the literature \cite{vainstein2007does, wang2012if}.
%With this action set, we also note a novel result, which is that agents who pick \textit{copy-the-best} form the majority of cooperators and cooperation clusters, while defectors usually are the ones choosing to move.
%\textcolor{orange}{RESULTADO? Having seen that our results with this action are similar to literature and also obtaining a new result, we showcase the ability of the algorithm to model game-theoretical scenarios easily by performing a simple change, which is made by }
Then, to add population diversity to our study, we introduce %a last action to the previous action set, giving players the ability to   
 \textit{persist}, 
%This introduction of 
 a very distinct action that can be also viewed as allowing more neural diversity to be present, which was recently argued as being key to learning agents' success~\cite{bettini2024neural}.
% The use of a Fermi update together with the independent Q-learning algorithm was recently studied in a filled lattice \cite{(fermi-update-q-learning)guo2022effect}, but the essence of the simulation is differs from our approach.
%\st{In our model, to create this new set of actions and thus the new, more diverse simulation framework,} we
% To perform this change in the action set, we append a \textit{persist} action to the previous action set.
This action will, when taken, make the agent a static player, who will neither try to update its payoff nor to move.
% eu acho que essa parte aqui da pra manter
The simple addition of this seemingly simple action results in a striking symbiotic behaviour between agents that \textit{persist} and agents that \textit{copy-the-best}, causing cooperation to endure in settings where it before disappeared.
%\textcolor{orange}{RESULTADO? This simple change results in a striking symbiotic behaviour between agents that \textit{persist} and agents that \textit{copy-the-best}, causing cooperation to endure in settings it disappeared when the players could only compare their payoff.
%The behaviour is mutualistic, as we will argue in the text, as it is a net benefit for these agents, that in most cases are the ones forming strong cooperation clusters.}

%\textcolor{orange}{DISCUSSAO/CONCLUSAO Our work, then, highlights the suitability of independent reinforcement learning algorithms in modeling evolutionary dynamics, showing with it that mobility and dilution can influence the learning agents' behaviour greatly and produce high levels of cooperation between them in specific settings.
%We believe that such implementations can help understand coordination in reinforcement learning agents through cooperation in mobile or diffusive environments, and also help study and benchmark large populations in multi-agent settings, which is a problem in itself when considering only reinforcement learning theory and applications \cite{canese2021multi}.}

%colocar alguns resultados e motivaçao
We also note two aspects of the connections between evolutionary game theory and reinforcement learning. 
First, in the reinforcement learning setting, asynchronous updates would be classified as a \textit{off-policy}~\cite{uehara2022review} updates when considering the simulation as a whole and an \textit{offline}~\cite{levine2020offline, prudencio2023survey} update, when taking each round as a separate training step.
Second, we note that the use of players and agents is exchangeable in our context and is thus used as so throughout the paper, as the former is more suitable for game theoretical scenarios and the latter for reinforcement learning ones.
When we combine game theory and reinforcement learning, these concepts are in essence the same. 
The difference would be, if %there is 
 any, that \textit{agent} serves to describe an individual population with different possible actions, while \textit{player} better describes an individual that is part of one of many populations, where each population is defined through their strategy, that is in turn defined by one of the actions.
This is all tied together in the end by the population-policy equivalence \cite{soma2024bridging, bloembergen2015evolutionary}, which will be explained below. 

%\st{, which will be better explained and serve as a tool for a more precise analysis of the last result of this paper.}
%%%%%
%%%%%
%%%%%
\section{\label{sec:model}Model}

In our model, the vertices of an $L \times L$ square lattice with periodic boundary conditions may be either empty or occupied by players. %In light of this, we consider spatial games on a square lattice occupied by 
%\textcolor{red}{\st{Throughout a simulation, the density of players, $\rho$, is kept fixed, leaving a fraction $1 -\rho$ of empty spaces.} - REPETIDO abaixo, manter?}  
 At each round, we sample a player at random to play with its von Neumann neighbourhood \cite{toffoli1987cellular}; this  sampling is repeated $L^2$ times to complete a Monte Carlo Step (MCS).

For the game results, we use the rescaled payoffs $R=1$, $P=0$, $S=0$ and $T=b$, with $b\in(1, 2)$, in which the interval for $b$ is defined in order to preserve the inequalities that define the weak prisoners dilemma game, and characterizes the temptation to defect \cite{nowak1992evolutionary}. 
This defines a payoff matrix given by
\begin{equation}
 \mathbb{P} =
    \begin{bmatrix}
        1 & 0 \\
        b & 0
    \end{bmatrix}.
    \label{eq:payoff-matrix}
\end{equation}

Using matrix notation, the payoff at each sampling step is determined by the state of a player $k$ and its neighbourhood (we define $s_{k,C} = [1, 0]^T := C$ and $s_{k,D} = [0, 1]^T := D$)  as %(we define $s^k_C = [1, 0]^T := C$ and $s^k_D = [0, 1]^T := D$)  as
\begin{equation}
   \pi_k(t) =
      \sum\limits_{\langle ik\rangle}s_k^T \ \mathbb{P} \ s_i, 
   % \pi_k(t) =
   %   \sum\limits_{\langle ik\rangle}s_k \ \mathbb{P} \ s_i, 
    %P^k(t) =
    %  \sum\limits_{\langle ik\rangle}s^k \ \mathbb{P} \ s^i, 
      \label{eq:payoff}
\end{equation}
where the sum over $\langle ik\rangle$ represents an iteration through the nearest neighbours of player $k$. %, and the matrix $\mathbb{P}$ is given by (\ref{eq:payoff-matrix}). 

%falar mais sobre isso na introduçao, sobre os diferentes tipos de artigos que foram feitos no assunto
%Recent uses of the Q-learning framework for the study of spatial prisoner's dilemma games \cite{zhang2019collective, zhang2020oscillatory, wang2022levy} have only considered a static and filled environment, in which players are obliged to play with they neighbours at each step.
% pessoal chama de diluir a rede
% talvez tenha que dizer que passa a ter duas variaveis. a de ocupacao e a do jogo
In order to introduce defects and mobility to the already studied static reinforcement learning framework \cite{zhang2019collective, zhang2020oscillatory, wang2022levy}, we initially dilute the lattice with an exact number of defects, defining  a density $\rho\in(0, 1]$ of occupied sites.
%nao e melhor chamar de rho ? dai lembra densidade...
With this concept in place, we introduce the independent Q-learning algorithm model with the definition of the state and action spaces.
We separate our model into different case studies, all of which  use the same state set $\mathbf{s} = \{C, D\}$, defined as the strategy of the player in the previous round, and different action spaces $\mathbf{a_\square}$ that will be discussed below.

For the first studied case, we use the static action set $\mathbf{a_S} = \{C, D\}$, which defines simply the action to cooperate or defect in the round. 
With this, players can decide only which strategy they will use in the round, and know nothing about their surroundings, besides their own payoff.

Having studied the static case in a diluted lattice, we first introduce mobility, $M$, in the form of a diffusive movement, so that the new set of actions becomes $\mathbf{a_M} =\{C, D, M\}$. 
In a more precise definition, the novel action $M$, then, represents the decision to  randomly move to a vacant space in the neighbourhood instead of playing at the present location.  
%At the destination, the player  maintains its previous state and receives a payoff from its new neighbourhood.%, only then updating its Q-table. 
%In a more precise definition, the novel action $M$, then, represents the decision to not play at the present location and to randomly move to a vacant space in the neighbourhood, maintaining its previous state and receiving a payoff from its new neighbourhood. 
A mobility rate $p_d\in [0, 1]$ is employed, which will give a probability of diffusion success when moving, and thus allow us to analyse how cooperation is affected by different levels of mobility for a given density.
If the player's attempt to move fails or there are no vacant spaces, we skip the interaction and the player does not update their state. %neither the state nor the Q-table.
When movement is successful, the player is relocated to a new site, leaving an empty one behind, and a game is played in the new site, updating the player's payoff.
It is important to note that, for $p_d=0$, we recover the static case in which players cannot move.

Finally, the actions $C$ and $D$ are removed %-- unconditional C and unconditional D?
 and experiments with the also novel actions $B$, or \textit{copy-the-best}, and $P$, or \textit{persist}, are performed.
Inspired by \cite{nowak1992evolutionary, vainstein2001disordered, vainstein2007does}, a player that chooses $B$ will copy the strategy of the best player in its neighbourhood, i.e., 
the one with the greatest payoff. % in its neighbourhood. 
This will be used to define an action set $\mathbf{a_B} = \{B, M\}$. 
When the $P$ action is taken, the player will %\st{stay in place and} 
 not change its state or its position. 
%\st{This helps us} 
This defines the final action set, $a_{\mathbf{B-P}}=\{B, P, M\}$.
%sera que podiamos seprar em dois jogos? um seria aprender a jogar e o outro a sair de conf desfavoraveis

Given the fixed state set throughout the paper, each set of actions generates a Q-table for each player $k$, where the table element $Q_{ij}$  represents the Q-value for state $i$ and action $j$, defining thus, in general form:
\begin{equation}
\mathbf{Q} = \ \begin{blockarray}{ c c c c l}
\BAmulticolumn{1}{c}{a_1} & \BAmulticolumn{1}{c}{a_2} & \BAmulticolumn{1}{c}{\cdots} & \BAmulticolumn{1}{c}{a_n} \\
\begin{block}{[ c c c c ] l}
  Q_{Ca_1} & Q_{Ca_2} & \cdots & Q_{Ca_n} &\multirow{1}{*}{ $C$}\\
Q_{Da_1} & Q_{Da_2} & \cdots & Q_{Da_n} &\multirow{1}{*}{ $D$}\\
\end{block}
\end{blockarray} 
\label{table:q-learning}
\end{equation}
in which the states in the row represent the strategy of the player as a cooperator ($C$) or as a defector ($D$) in the previous round, and the actions possible at every round.

To guarantee that all states are sufficiently visited, a stochastic factor is added in the form of an epsilon-greedy algorithm \cite{sutton2018reinforcement}.
The predefined value of $\epsilon$ is then a probability that dictates how often the player will take an action at random.
These exploring steps allow the players to obtain payoffs from all actions, thus guaranteeing that the state and action spaces are sufficiently visited, which is a condition for convergence of the single agents version of algorithm~\cite{watkins1992q}. 
This, in turn, also determines the convergence of our independent multi-agent algorithm.
% o que significa ser bootstrapped?
%get (oneself or something) into or out of a situation using existing resources.
%"the company is bootstrapping itself out of a marred financial past"

% nao sei se aplica bem nessa situaçao, mas foi oq me veio na cabeça na hora

If the decision is not made at random, %happening 
 with probability $1-\epsilon$, the Q-table will be used.
In this case,  the player in a state $s$ will choose the action that corresponds to the maximum Q-value in its %the corresponding 
 state row $s$ in matrix (\ref{table:q-learning}), i.e., $\max\{Q_{s, a_1}, Q_{s, a_2}, ..., Q_{s, a_n}\}$. 

After the decision is made, the player's Q-table is updated according to:
\begin{equation}
Q_{s,a}(t+1) = (1-\alpha)Q_{s,a}(t) + \alpha(\pi(t) + \gamma \max{(Q_{s^\prime, a^\prime})})
    %Q(t+1) = (1-\alpha)Q(t) + \alpha(R + \gamma \max{(Q_{s^\prime, a^\prime})})
    \label{eq:update_q_value}
\end{equation}
% ja tinha aparecido um gammC e gammD antes...
where $\alpha\in (0, 1]$ is the learning rate, $\gamma$ is the discount factor, which determines how much we want to consider possible future decisions into the update, and $\max({Q_{s^\prime, a^\prime})}$ is the maximum value in the Q-table associated with 
%the state in the next round $s^\prime$ and the future action $a^\prime$, all of which are determined by the chosen action $a$. 
the future state $s^\prime$ and action $a^\prime$ pair, which are determined by the chosen action $a$. 
The reward %$R$ 
 is considered to be equal to the player's payoff, $\pi(t)$, determined by equation (\ref{eq:payoff}).

In order to clarify the learning process, we summarize it in the following steps~\footnote{All code is available on \href{https://github.com/gustavomangold/}{github.com/gustavomangold/}.}: 
\begin{enumerate}
    \item We initialize an $L\times L$ square lattice by populating it partially, with a density $\rho$ of players occupying it randomly.
    \item Each player is randomly assigned its first role, or state, as cooperator or defector and their Q-table is initialized to zeros.
    \item A player is sampled at random, choosing its action either randomly with probability $\epsilon$ or according to the maximum value on the Q-table, with probability $1-\epsilon$, to obtain a payoff.
    \item The Q-table is then updated according to Eq.~(\ref{eq:update_q_value}), and the state of the player is updated based on the action $a$. 
    %\hmf{esse é o passo que sempre acho meio confuso..}
    \item A Monte Carlo Step (MCS) consists of $L^2$ repetitions of items $3$ and $4$ to complete a learning episode. 
    %or \hmf{acho que consistindo de ficaria melhor..} 
\end{enumerate}

\begin{comment}
Steps 3 and 4 can be further exemplified considering a situation where the sampled player is a cooperator and chooses to defect on the next round, the value then updated in the Q-table will be $Q_{C, D}$.  
The $\max{(Q_{s^\prime, a^\prime})}$ term is obtained by taking the maximum value possible for its next round state of defector, e.g., the maximum of the second row of (\ref{eq:q-table-no-knowledge-dynamic}).
\end{comment}

The game is iterated through a maximum of $N=10^5$ %decidir o N depois
steps on a $100 \times 100$ lattice to complete an asynchronous, single-agent update, Monte Carlo simulation;  simulations are evolved for at least $2\times 10^4$ MCS until the system reaches a steady state. We also carry out a minimum of $10$ and a maximum of $50$ independent experiments for statistical significance, with the presented results being the average value of the last $\sfrac{N}{10}$ steps of all said experiments. 
%\st{, which is where the system attains stationarity for all parameters.} 
The simulation parameters are kept fixed during each simulation and all other learning parameters are specified in each section.

%%%%%
%%%%%
%%%%%
\section{\label{sec:results}Results and discussion}
%\st{In this section, several simulations are conducted for a number of different parameters.} 
We divide the results into two cases with two posterior subdivisions, first studying dilution in the no-knowledge case, then introducing mobility and surrounding knowledge with new action sets.
At each section, a set of learning parameters is specified and chosen specifically for convergence purposes.
%For both cases, we first observe how the learning parameters from equation (\ref{eq:update_q_value}) influence the cooperation and the errors, in order to find an optimal setting for our experiments.
%fiquei um pouco em duvida sobre o que sao os settings
%talvez separar assim a seçao?
\subsection{{\label{sec:results-no_knowledge} No-knowledge case}}
We begin by analysing the case where the actions do not take into account any neighbourhood information in the algorithm except for the obtained payoff.
In this setting, each player's actions are limited to changing their own state.
%\st{This subsection uses the learning parameters} 
 Here $\epsilon=0.02$, while we set $\gamma=0.8$ and $\alpha=0.75$, with $N=2\times 10^4$ total steps run for $10$ independent configurations.
 % robustez dos parametros, muda qualitativamente?  mudar gamma e alpha tem algum limit que funcione ou não?
\subsubsection{{\label{sec:results-static}} Static agents}
\begin{figure}[t]
\setkeys{Gin}{width=\columnwidth}
    \centering
    \begin{subfigure}{0.95\columnwidth}
     \includegraphics{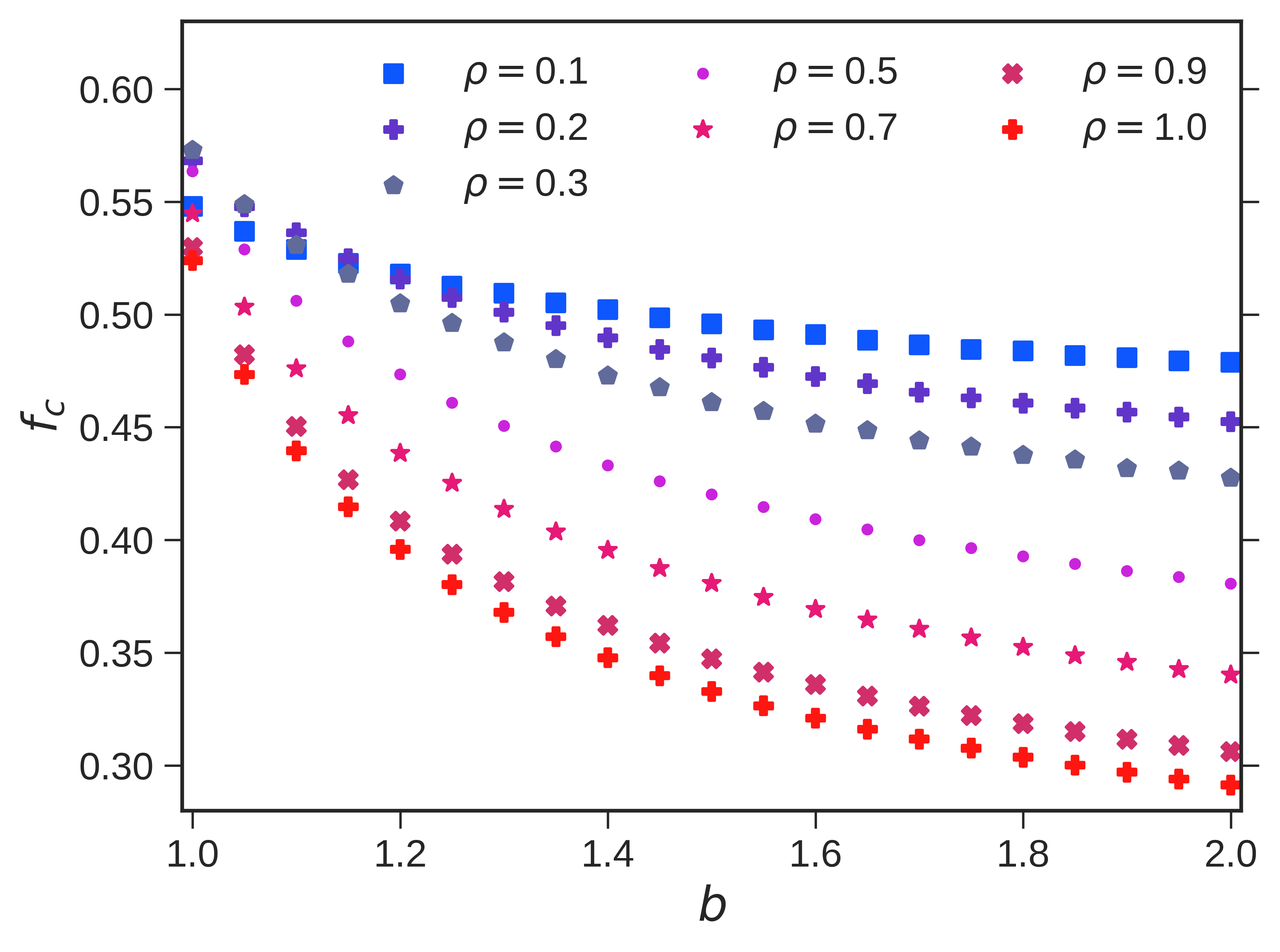}  
    \end{subfigure}\hfill
\caption{%\textcolor{red}{Aumentar tamanho números } 
Cooperation versus the temptation to defect for several densities with static agents, showing the effect of dilution in the lattice.}
\label{fig:coop-versus-b-action-c-or-d} 
\end{figure}

With no mobility, that is, in the diluted setting with an exact number of holes, we use the first set of actions $\mathbf{a_S}$ as described in Section \ref{sec:model}, defining thus the Q-table:
\begin{equation}
\mathbf{Q_s} = \ \begin{blockarray}{ c c }
\begin{block}{[ c c ]}
  Q_{CC} & Q_{CD} \\
Q_{DC} & Q_{DD}\\
\end{block}
\end{blockarray}.
\label{table:q-table-c-or-d-static}
\end{equation}
%\st{The results of the simulations are shown in} In this case, in Fig.~\ref{fig:coop-versus-b-action-c-or-d}, we see a monotonic rise in cooperation levels with decreasing  density $\rho$, and that the fraction of cooperators $f_C$ becomes weakly dependent on the temptation to defect $b$ for low densities, which might seem counter intuitive.
In Fig.~\ref{fig:coop-versus-b-action-c-or-d}, we observe a monotonic increase in cooperation levels as the density $\rho$ decreases. At low densities, the fraction of cooperators $f_C$ becomes weakly dependent on the temptation to defect $b$, which may seem counterintuitive.
However, this can be easily explained: at low densities, such as $\rho = 0.1$, players are more likely to be isolated and lack neighbors, leading them to choose their actions randomly -- essentially flipping a coin at each round. In this scenario, we naturally expect a balanced mix of cooperators and defectors. 
%However, this is easily explained for low densities such as $\rho=0.1$, as players are more likely to be isolated and thus without neighbours, deciding which action to take with a coin flip at each round. 
%In this situation, we of course have an expected equal mixed state of cooperators and defectors.
% talvez falar de cluster? percolaçao?
For higher occupations, such as $\rho=0.5$, the aforementioned isolation still takes place, but to a lesser degree, still being able to influence in cooperation levels.

Here, two points are worthy of mention. First, the cooperation levels observed for a fully occupied lattice are consistent with previous findings using the same Q-learning framework~\cite{wang2022levy, yang2024interaction}. 
Second, we observe that across all densities, cooperation does not fall to zero as the temptation to defect, $b$, increases. This stands in stark contrast to the spatial PD played with fixed update rules, such as those based on the Fermi-Dirac transition probabilities, where cooperators rapidly go extinct under similar conditions~\cite{perc2008social}.
%This difference can be explained by the way states and actions are defined in the Q-learning algorithm: players learn that if everyone defects -- even under high temptation, as shown in Fig.~\ref{fig:coop-versus-b-action-c-or-d} -- no player receives any reward.
%As a result, the system naturally stabilizes in a mixed state of cooperation and defection.
%concerns an observation that for all densities in this setting cooperation does not drop to zero as $b$, the temptation to defect, grows.
%That is polar opposite to the spatial prisoner's dilemma played only with fixed update rules, such as with a Fermi-Dirac distribution, where we see cooperators quickly becoming extinct in the same regime~\cite{perc2008social}.
This fact is explained by noting that with this definition of states and actions in the Q-learning algorithm, players learn that if everyone turns to defection, even under high temptation as seen in Fig.~\ref{fig:coop-versus-b-action-c-or-d}, no player receives any reward. As a result the system  stabilizes in a mixed state of cooperation and defection. %and therefore there will always be a mixed state of cooperation and defection. 
%aqui eu poderia explicar o porquê disso (eu sei por que), mas nao sei se cabe? ver mais tarde
%talvez citar alguma ref sobre assimetria/nao reciprocidade e cooperaçao, se tiver
%That is, a defector can have three neighbours, while the cooperator has only one
\subsubsection{{\label{sec:results-dynamic}Diffusing agents}}
\begin{figure}[tbhp]
\setkeys{Gin}{width=\columnwidth}
\captionsetup[subfigure]{justification=centering}
    \centering
    \begin{subfigure}{0.5\columnwidth}
    %\begin{subfigure}{0.499\textwidth}
     \includegraphics{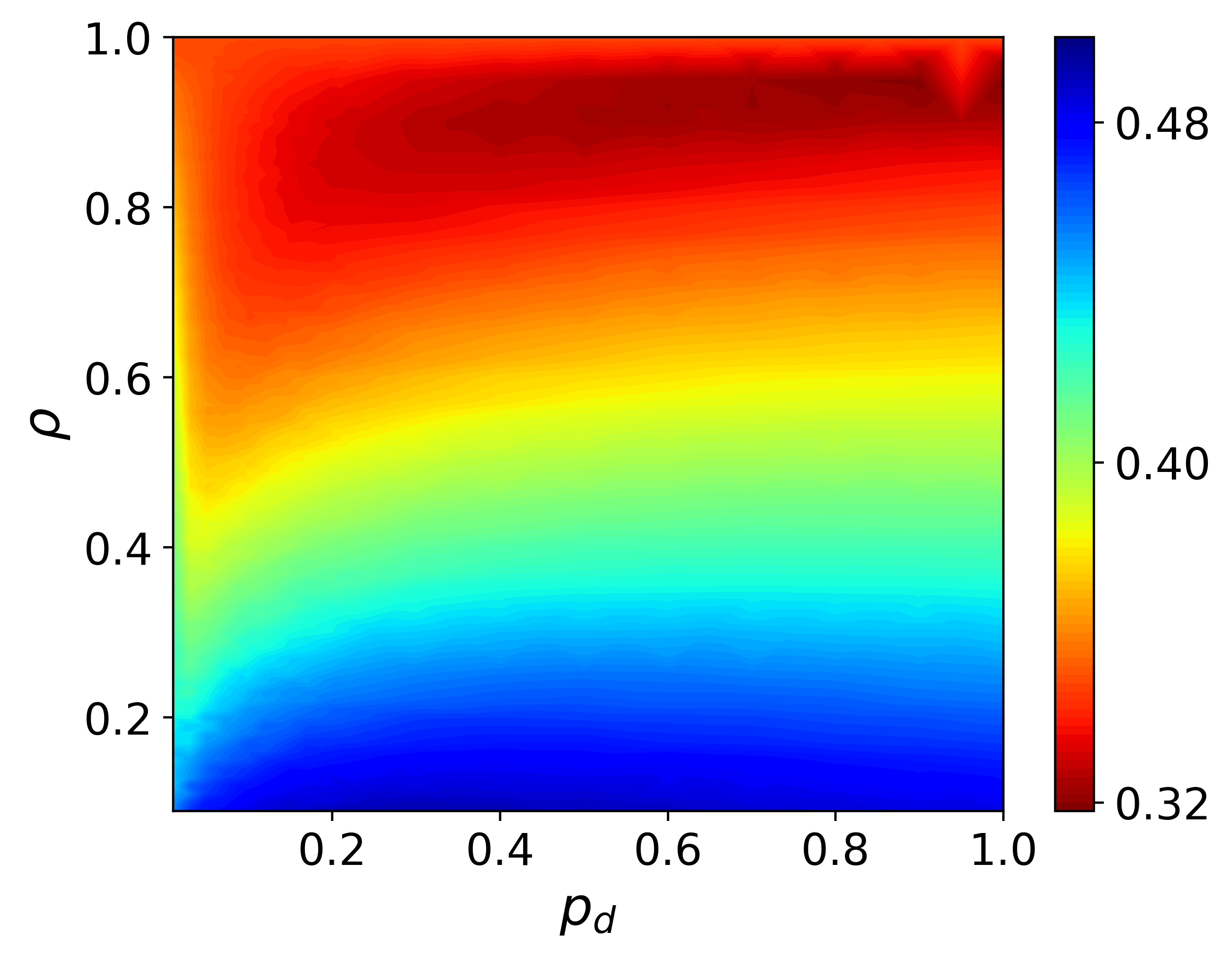}  
     \caption{}
     \label{fig:heatmap-coop-mobility-action-c-or-d} 
    \end{subfigure}\hfill
    \begin{subfigure}{0.5\columnwidth}
    %\begin{subfigure}{0.499\textwidth}
     \includegraphics{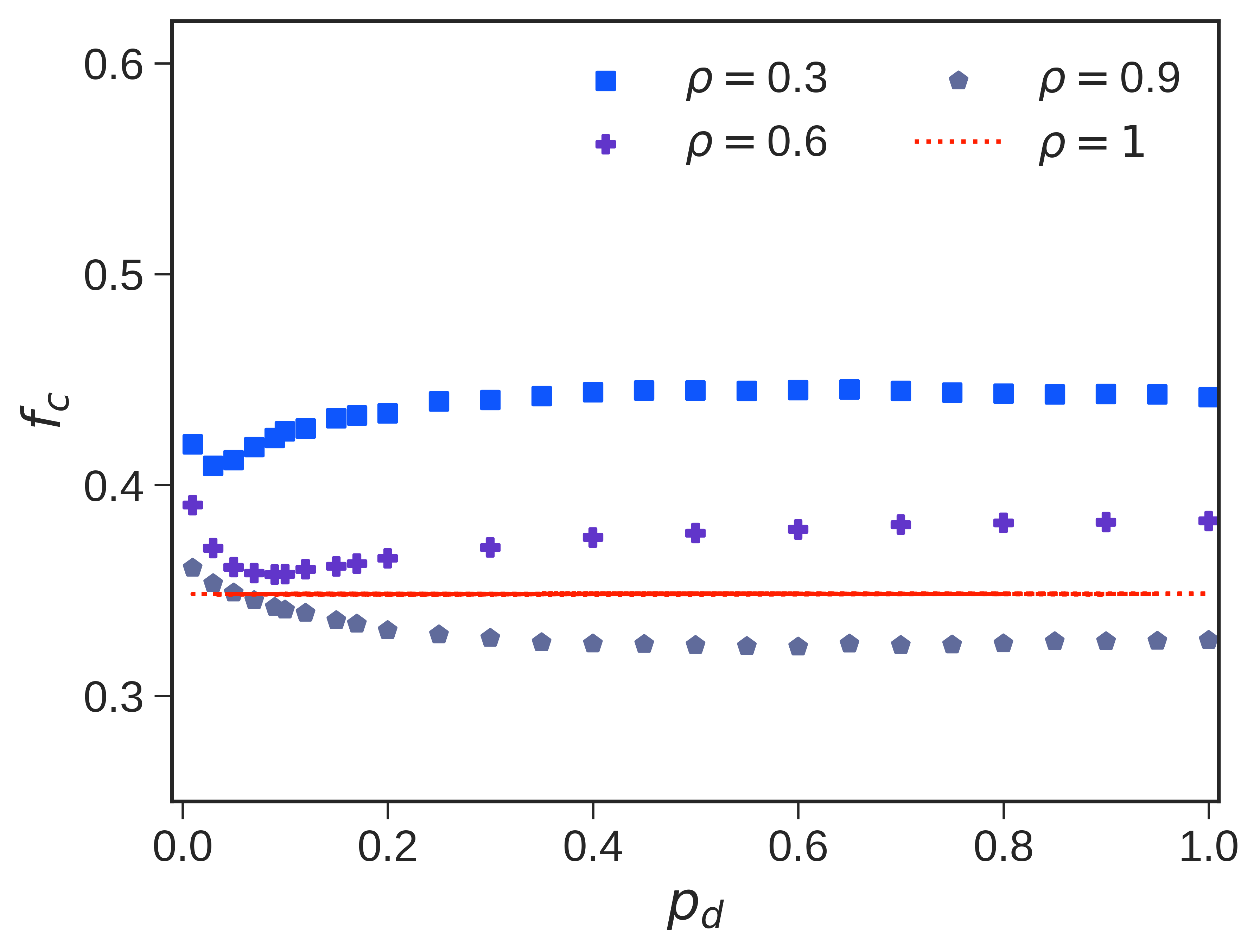}  
     \caption{}
     \label{fig:mobility-curves-action-c-or-d} 
    \end{subfigure}\hfill
\caption{%\textcolor{red}{números pequenos nas figuras - Fica ruim se  escala de cor for de 0 a 1? Para ficar mais fácil de comparar com as outras figuras. Mudar cor do rho=1 na b.} 
Cooperation as a function of mobility and occupation density for the no-knowledge case with set $\mathbf{a_M}$ for $b=1.4$. \textbf{(a)} Heat map as a function of the density and the mobility rate, where the color bar shows the fraction of cooperators, which lies only in a small regime. \textbf{(b)} Curves for specific densities, showcasing the weak dependence of cooperation on mobility for this set of actions.}
\end{figure}

After considering the static case, the natural step is to introduce mobility, and we do that with the set $\mathbf{a_M}$, generating the Q-table:
\begin{equation}
\mathbf{Q_M} = \ \begin{blockarray}{ c c c}
\begin{block}{[ c c c]}
  Q_{CC} & Q_{CD} & Q_{CM} \\
Q_{DC} & Q_{DD} & Q_{DM}  \\
\end{block}
\end{blockarray}.
\label{table:q-table-c-or-d-dynamic}
\end{equation}
Results are shown in the heat map in Fig.~\ref{fig:heatmap-coop-mobility-action-c-or-d}, which shows that cooperation still increases when the system is diluted, as the previous case, but decreases when agents are mobile.
This behaviour is also shown for specific densities in Fig.~\ref{fig:mobility-curves-action-c-or-d}, where we can see that cooperation varies in a strict regime, as can be seen by the range of the color bar in the heat map.
 Furthermore, we see two distinct regimes: first, the lowest possible cooperation levels are attained in the limiting case of $\rho \approx 1$, the almost filled lattice. Second, for all lower densities shown, the lowest level of cooperation is seen for low mobility $p_d \approx 0$.

However, both cases can be explained by the same underlying behaviour. When agents attempt to diffuse in the $p_d \approx 0$ regime, movement is highly unlikely due to low probability. Similarly, in the $\rho \approx 1$ case, the lattice is nearly full, leaving little room for movement, and diffusion is again severely limited.
% efeito do p_d segue sendo notado no fc (?)
In both scenarios, agents that choose to move will likely remain stationary. Moreover, without any knowledge of their surroundings, clustering becomes difficult and 
if these stationary agents happen to be cooperators, they are easily targeted by defectors.
In the context of reinforcement learning, this dynamic reinforces defection, since being a defector leads to higher individual rewards, the action $D$ is positively reinforced, ultimately leading to a decline in cooperation.
%However, both cases can be explained by the %fall under the 
 %same general behaviour. 
%When agents try to move in the $p_d \approx 0$ regime, they most probably will not, for diffusion is very difficult. 
%In the $\rho \approx 1$ case, the lattice is almost without room for movement, so diffusion is very difficult as well. 
%That means that agents who choose to move, in these regimes, will probably just stay still.
%Paired with that, no surrounding knowledge is present, so clustering becomes very hard and, if these still agents are cooperators, they will be easily preyed on by defectors.
%With reinforcement learning, this means that the defection behaviour will be reinforced, as being one will increase the reward of an individual and the action $D$ will be reinforced, decreasing cooperation.

The most important takeaway %, however,
 is that in the absence of knowledge of the neighbourhood, low mobility causes the lowest cooperation levels.
This fact will be important when contrasting with the situation where agents have knowledge about their surroundings.
\subsection{{\label{sec:exploring-actions}} Exploring different actions}
Having analyzed the effect of mobile agents in the no-knowledge case, we turn to novel actions that take into account neighbourhood information.
This is based both on the literature of the spatial prisoner's dilemma \cite{nowak1992evolutionary,vainstein2007does} and on the reinforcement learning one, as players have shown to increase their learning rate when more information is available about their surroundings \cite{tan1993multi}.
Although more knowledge is now present, the algorithm is still an independent multi-agent framework, as each player has its own independent Q-table. % s are set to every player independently.
%For this subsection, a parameter change takes place, with 
In this subsection, we use a total number of steps $N=10^5$ run for $20$ independent samples, $\alpha=0.75$, $\gamma=0.8$, $\epsilon=0.15$. 
% Choose epsilon for convergence purposes, b=1.4 talvez da pra por algo pq eh numa das transiçoes la
%\textcolor{red}{Por que se usa um epsilon maior? Algo a dizer sobre b=1.4?}
Also, we fix the temptation at $b=1.4$ for all samples, as a transition in cooperation clusters with similar rule sets appears around this limit \cite{nowak1992evolutionary}.  % colocar algo sobre as transições e citar (nowak talvez)
\subsubsection{{\label{sec:greatest}} Copy-the-best}
\begin{figure*}[t!]
\setkeys{Gin}{height=.75\textwidth}
\captionsetup[subfigure]{justification=centering}
    \centering
    \begin{subfigure}{0.499\textwidth}
     \includegraphics{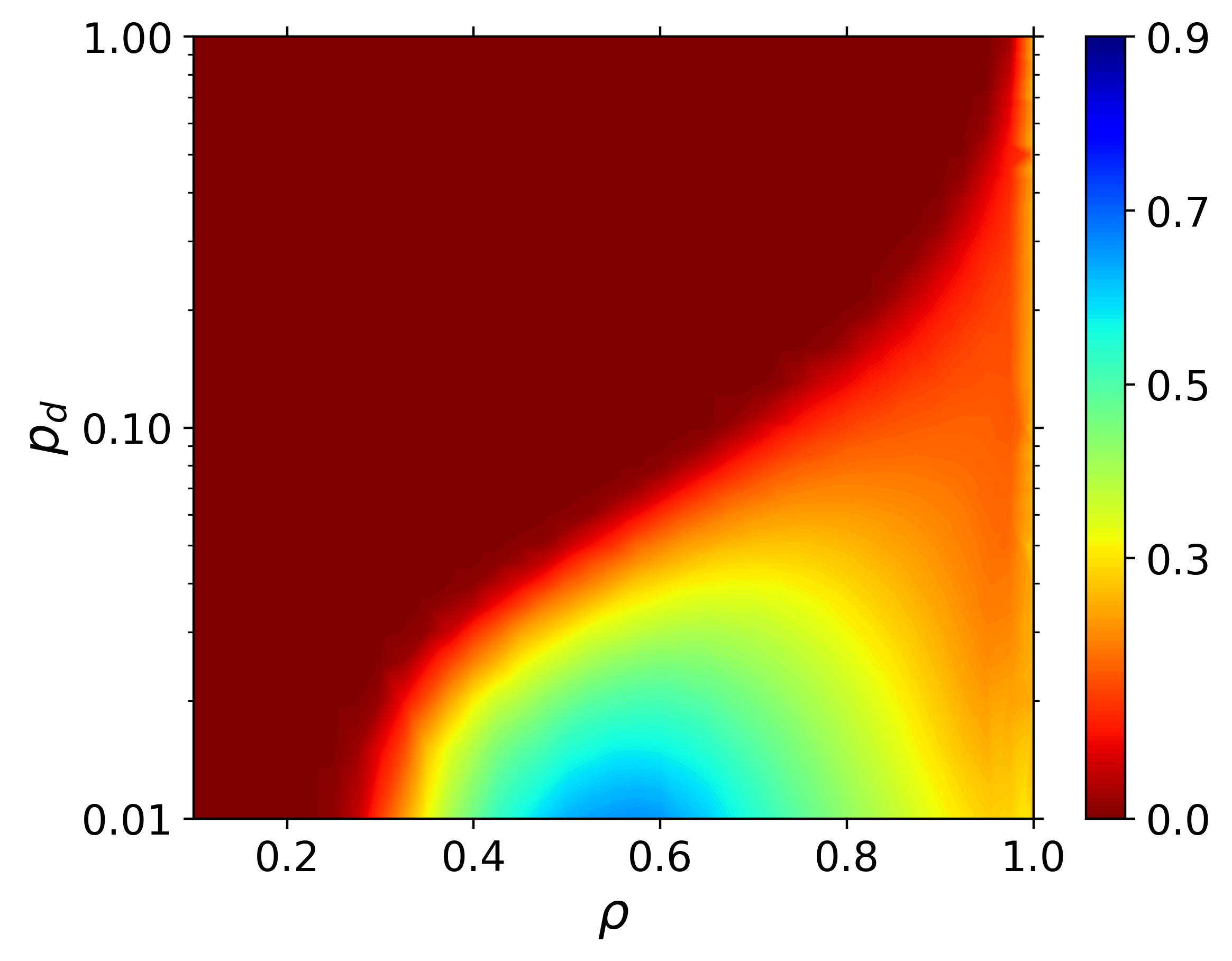} 
     \caption{}
     \label{fig:greatest-heatmap}
    \end{subfigure}\hfill
    \begin{subfigure}{0.499\textwidth}
     \includegraphics{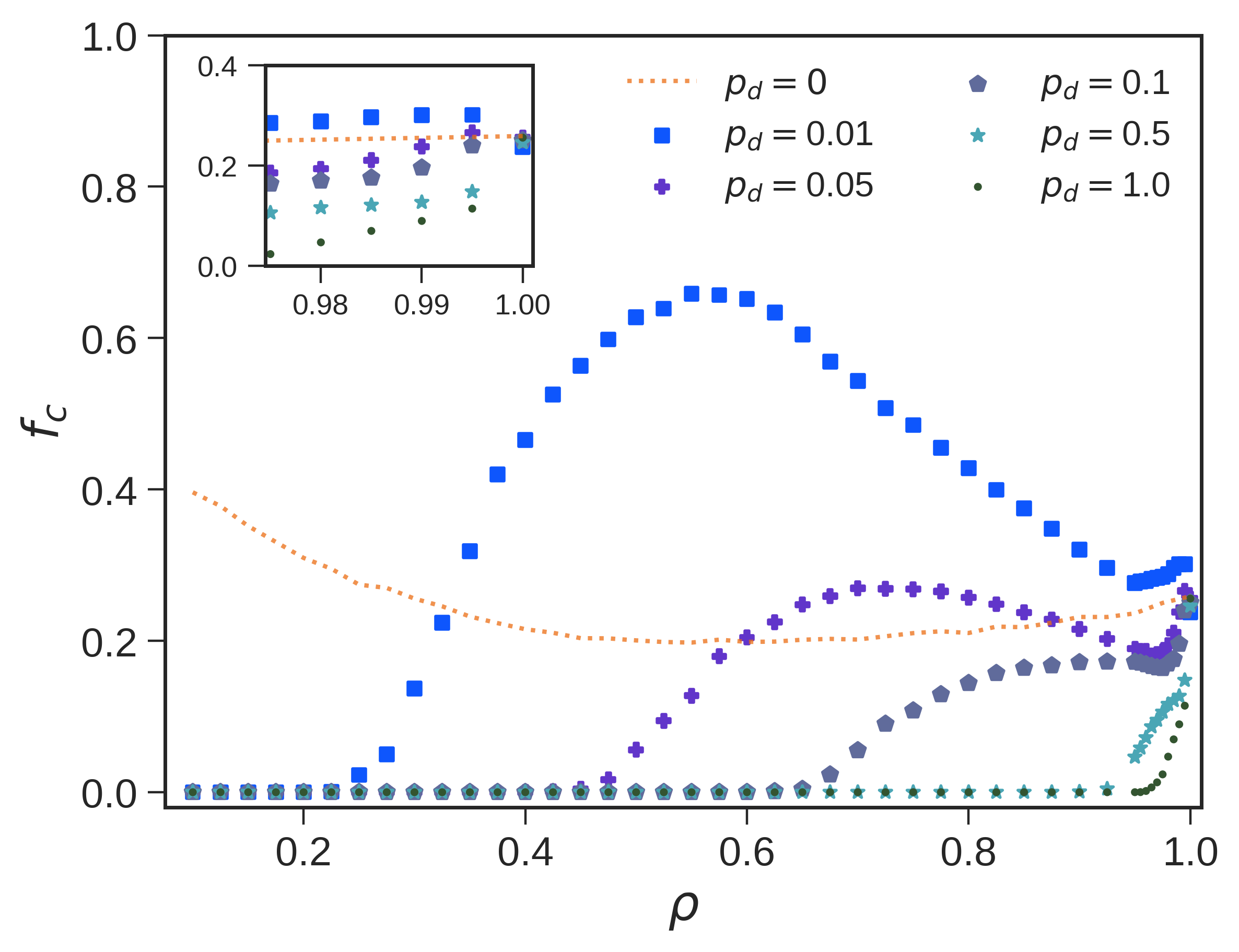}
     \caption{}
     \label{fig:greatest-coop-curve}
    \end{subfigure}\hfill
\caption{Fraction of cooperators as a function of the density of occupation and mobility with choosing the best player set $\mathbf{a_B}$, showing that there is an extensive region of null cooperation for almost all parameters, with a cooperation peak for low mobility around the percolation threshold. \textbf{(a)} Heat map, where the color bar represents the fraction of cooperators and the $p_d$ axis is in logarithmic scale. \textbf{(b)} Cooperation curves for different rates of mobility, showing peaks around the percolation threshold of $\rho\approx 0.6$ when mobility is decreased and agents arrange in cooperation clusters.}
\label{fig:greatest-coop-plots}
\end{figure*}

\begin{figure}[bhtp]
\setkeys{Gin}{height=1.\columnwidth}
    \centering
    \begin{subfigure}{0.32\columnwidth}
     \includegraphics{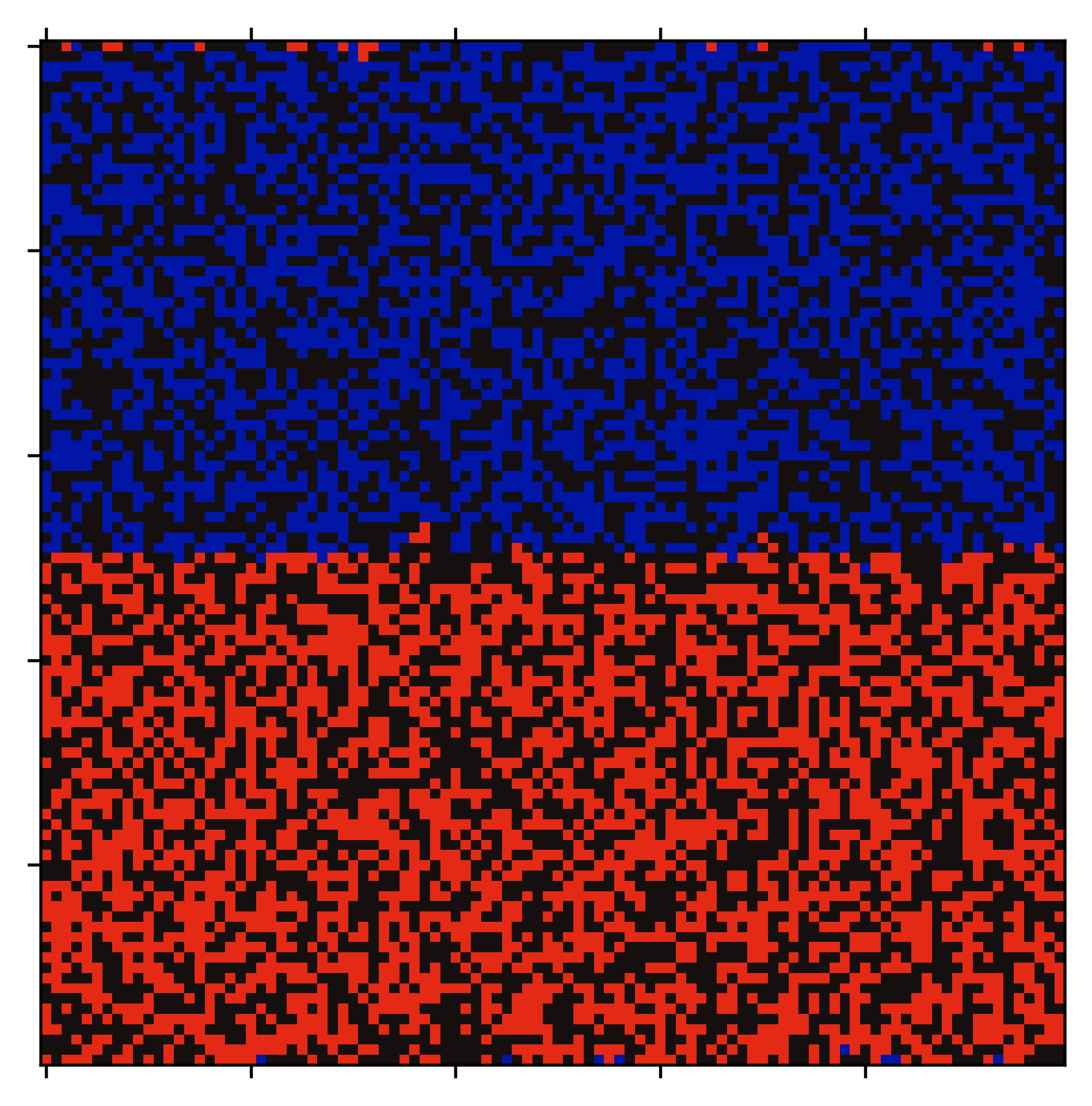} 
    \end{subfigure}\hfill
    \begin{subfigure}{0.32\columnwidth}
     \includegraphics{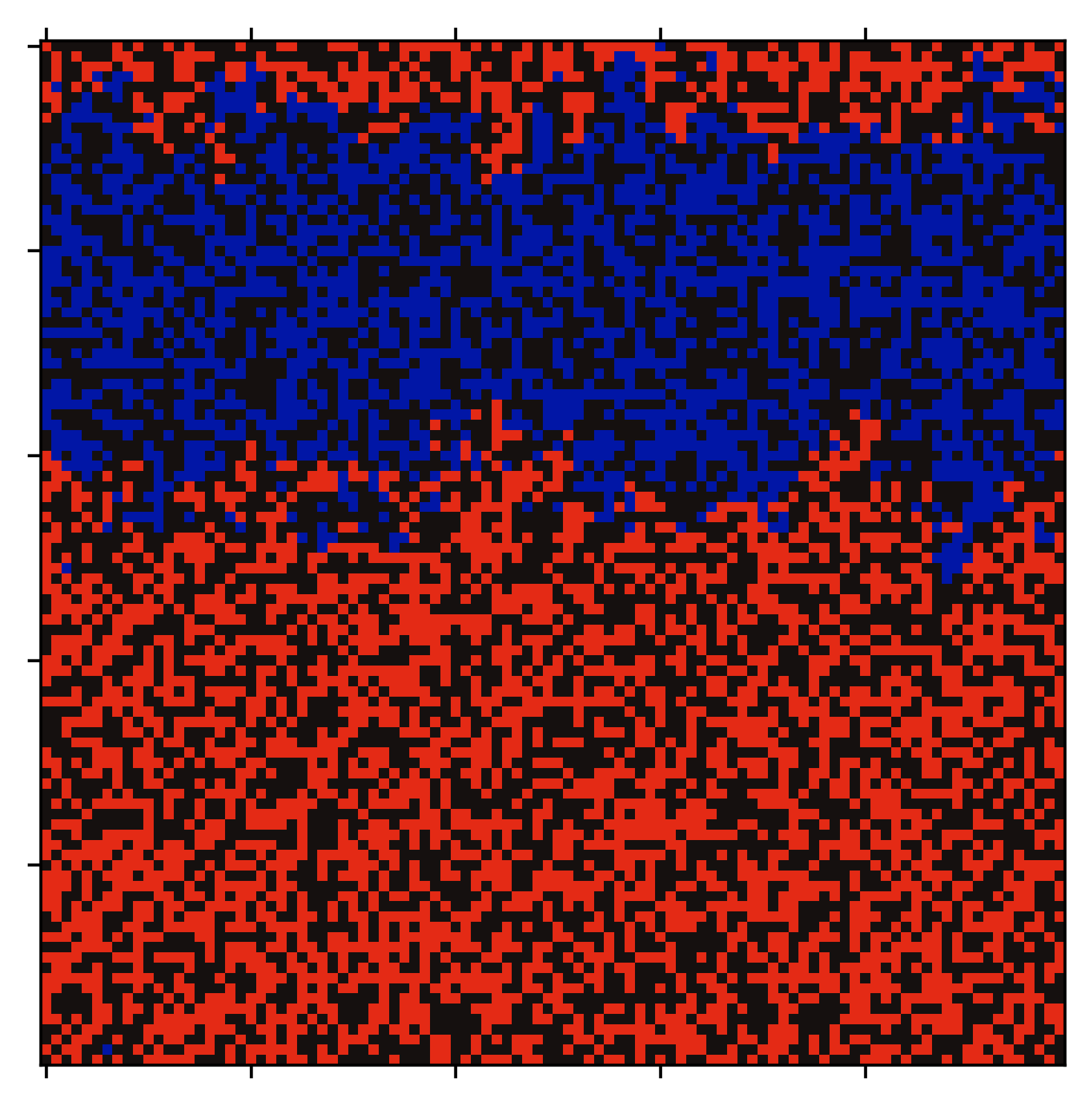}
    \end{subfigure}\hfill
    \begin{subfigure}{0.32\columnwidth}
     \includegraphics{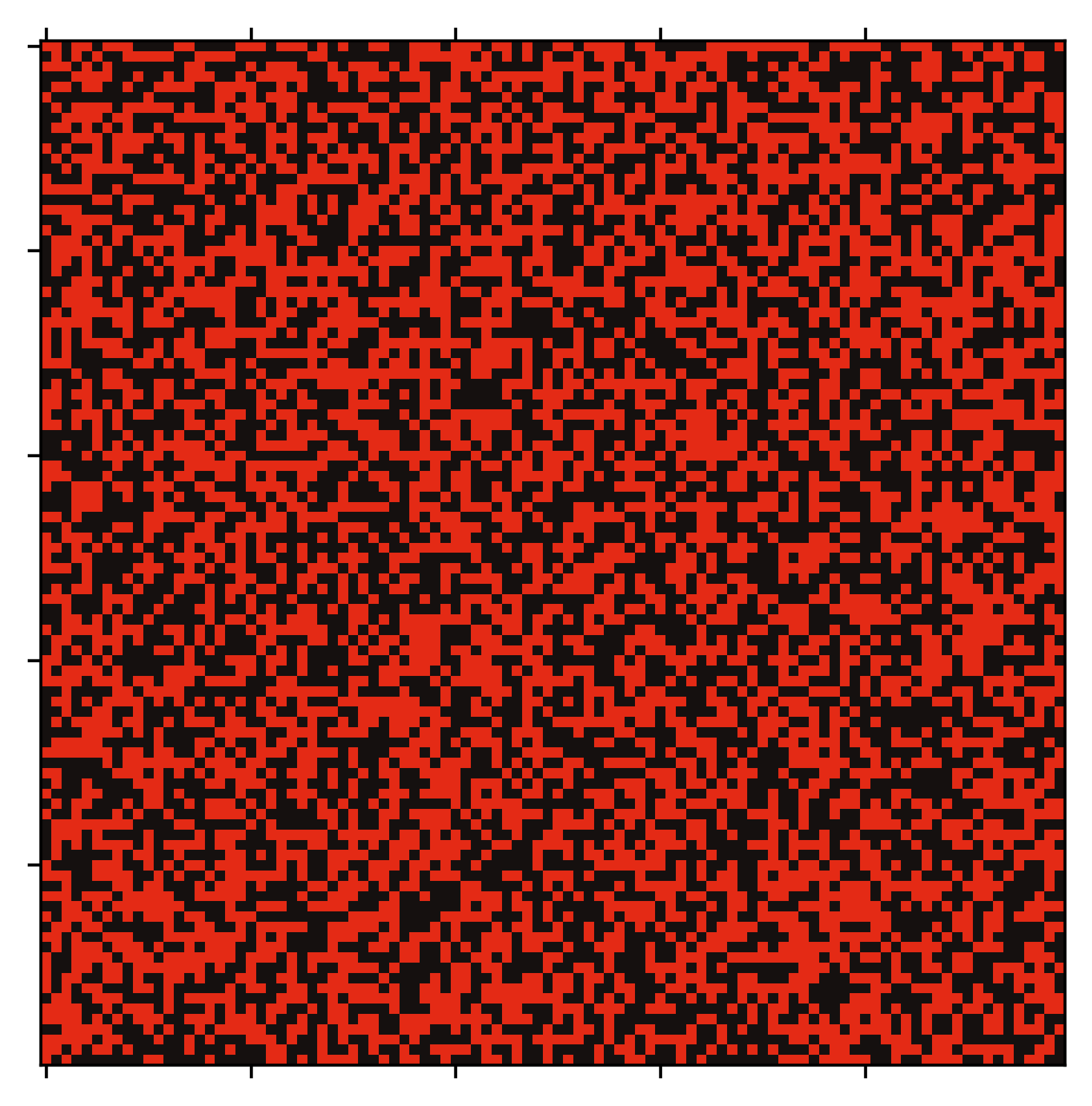} 
    \end{subfigure}\hfill
    \\
        \begin{subfigure}{0.32\columnwidth}
     \includegraphics{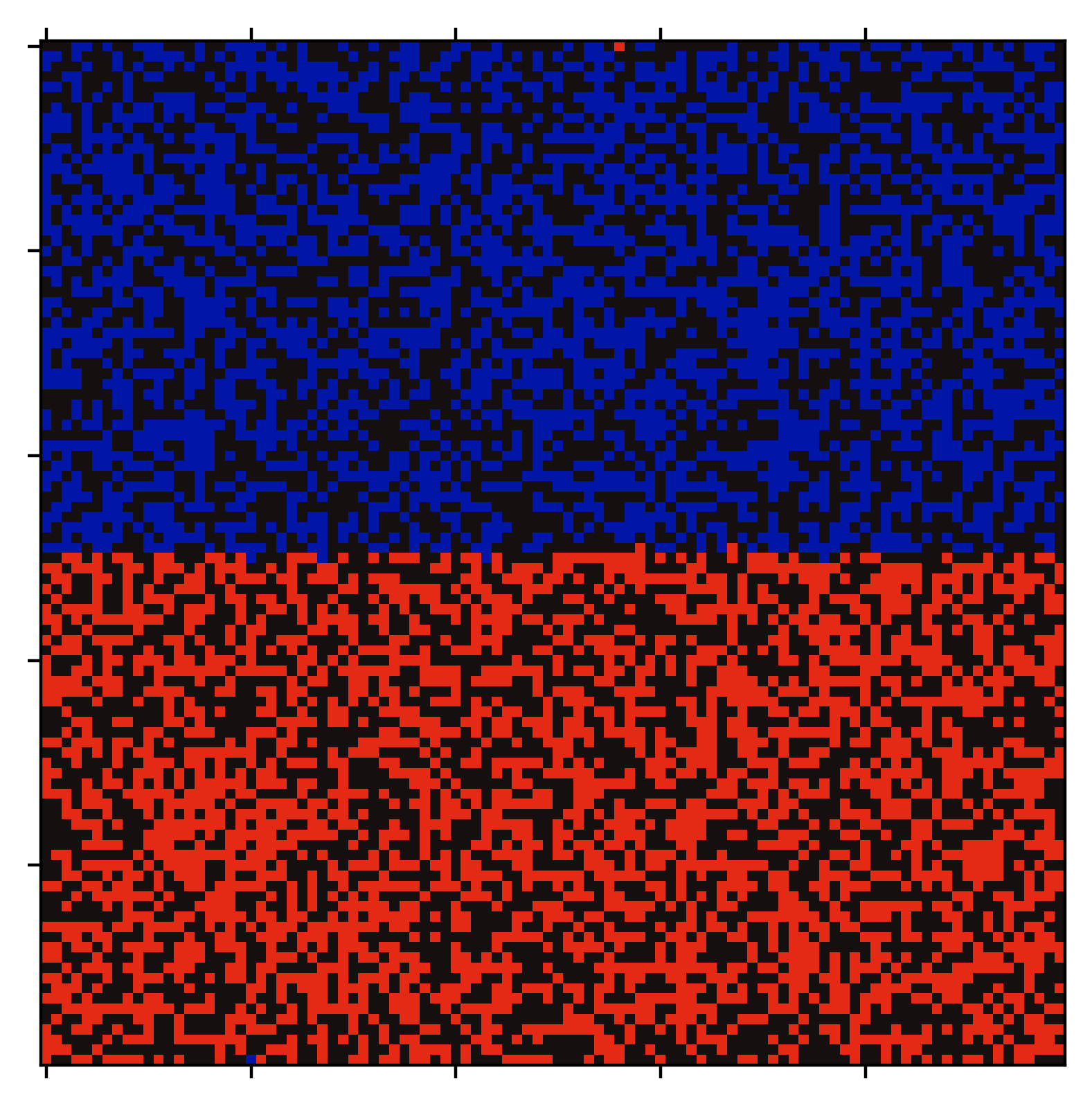} 
    \end{subfigure}\hfill
    \begin{subfigure}{0.32\columnwidth}
     \includegraphics{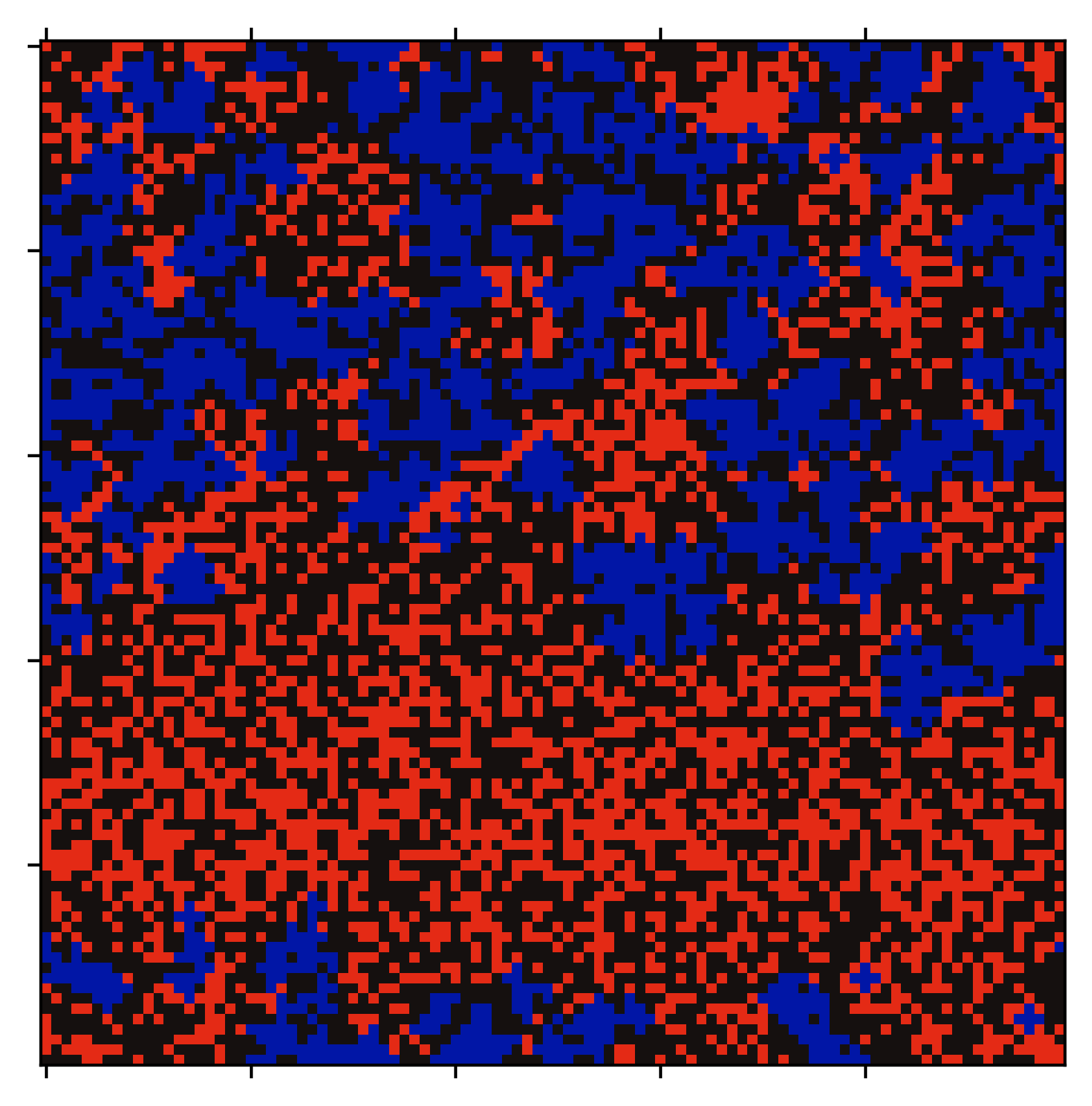}
    \end{subfigure}\hfill
    \begin{subfigure}{0.32\columnwidth}
     \includegraphics{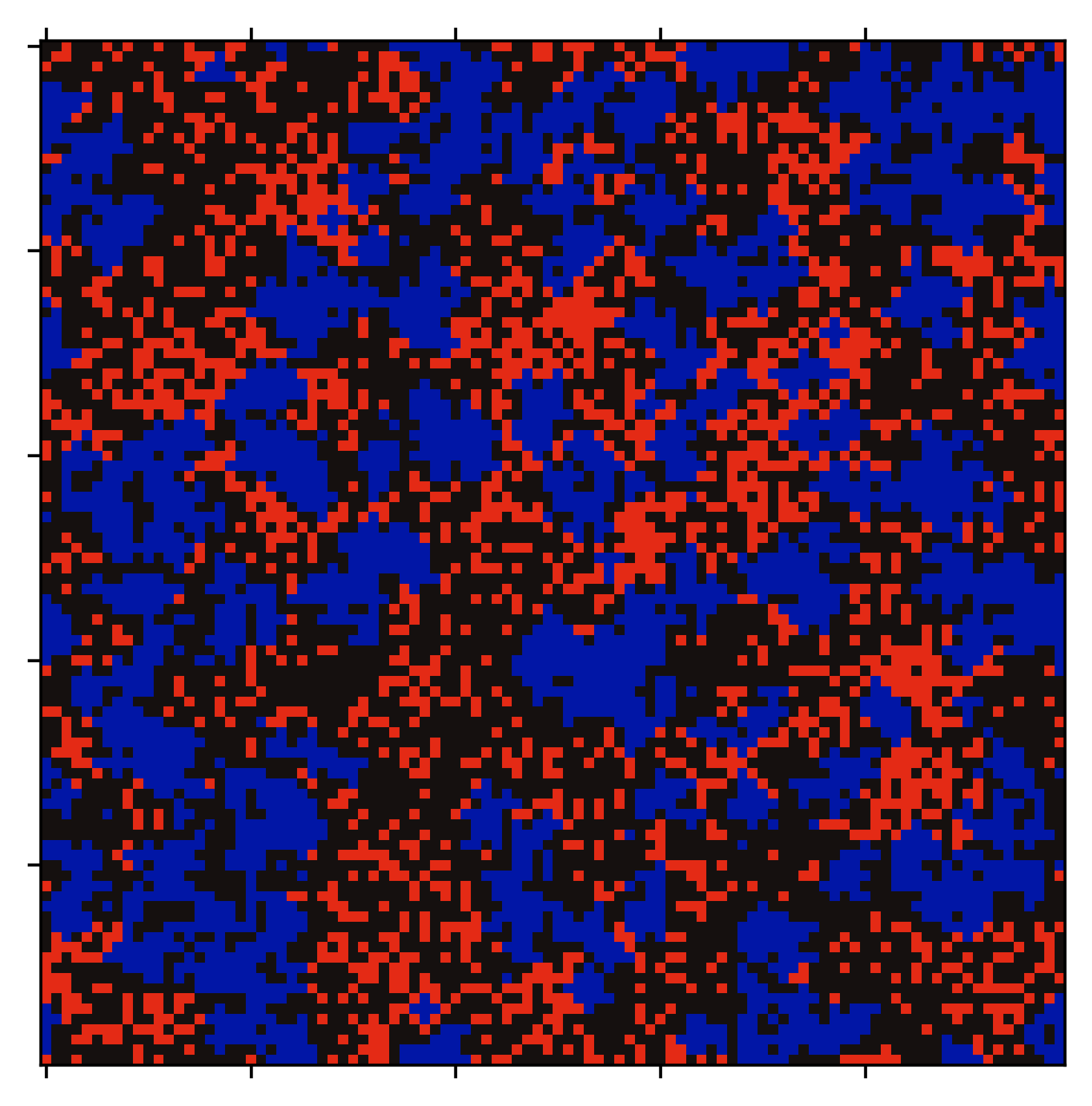} 
    \end{subfigure}\hfill
\caption{Snapshots with a specific striped initial configuration, taken at the percolation limit $\rho\approx 0.593$. Time increases to the right with different scales, with the last snapshot in the upper row at step 
$110$ and in the bottom row at step $10^5$ representing, respectively, $p_d=1$ and $p_d=0.01$, showing that fast agents tend toward defection quickly, while slow agents can gather in clusters and resist invasion from defectors.}
% falar sobre flutuações, ser estático ou dinamico os clusters?
\label{fig:snaps-greatest}

\end{figure}

We begin by simulating agents using the action set $\mathbf{a_B} = \{B, M\}$, which explicitly provides them with information about their surroundings.
This is achieved through the ability to choose an action that reveals the identity of the best-performing player nearby. 
This action set results in the following Q-table:
%We first simulate agents using the action set $\mathbf{a_B} = \{B, M\}$, which makes them have implicit knowledge of their surroundings, given through their ability to choose an action that will give them information about who the best player in their surroundings is. 
%This action set produces the Q-table:
\begin{equation}
\mathbf{Q_B} = \ \begin{blockarray}{ c c}
\begin{block}{[ c c]}
  Q_{CB} & Q_{CM} \\
Q_{DB} & Q_{DM}  \\
\end{block}
\end{blockarray} \ .
\label{table:greatest}
\end{equation}
%\st{When running experiments in which each agent has access to a copy of the Q-table,} 
 In this case,  we observe that cooperation is completely suppressed across almost all densities and mobility rates, as shown in Fig.~\ref{fig:greatest-heatmap}. It is interesting to note, %by looking at the curve for the static agents with $p_d=0$ 
  in Fig.~\ref{fig:greatest-coop-curve}, that there is a discontinuous transition from the static ($p_d=0$) to the low mobility regime that increases cooperation for regions of density around the site percolation threshold of the square lattice ($\rho_p \approx 0.593$ ~\cite{newman2000efficient}), while decreasing it for lower densities. %except in the region at high densities and, more interestingly, near the site percolation threshold of the square lattice ($\rho_p \approx 0.593$ ~\cite{newman2000efficient}). 
   A closer examination of this regime %, presented in Fig.~\ref{fig:greatest-coop-curve}, 
    reveals a peak in cooperation that shifts towards this threshold as agents become slower, what allows cooperators to cluster together and resist defector invasions. This clustering effect is illustrated in the snapshots of Fig.~\ref{fig:snaps-greatest} under low mobility conditions, in contrast to the scenario for $p_d \rightarrow 1$, where defectors successfully invade cooperative clusters.

%Running experiments with a copy of this Q-table available to each agent, we see that cooperation vanishes for almost all densities and mobility rates when looking at Fig.~\ref{fig:greatest-heatmap}, \textcolor{red}{FIG 5 estava citada antes da 3 -- mudei a ordem} except for a region in the higher density limit, and, more interestingly, for densities near the percolation limit of the square lattice, evaluated at $\rho \approx 0.593$ \cite{newman2000efficient}.
%Looking closer into this regime as shown in Fig.~\ref{fig:greatest-coop-curve}, we see a peak in cooperation that appears to move closer to this same limit as agents become slower, and thus cooperators are able to cluster together and resist. 
%This clustering is shown by the snapshots in Fig.~\ref{fig:snaps-greatest} for low mobility, contrary to the depiction for $p_d \rightarrow 1$ of the successful invasion of defectors onto the cooperation cluster.

\begin{figure}[thbp]
    \centering
    \includegraphics[width = .96\columnwidth]{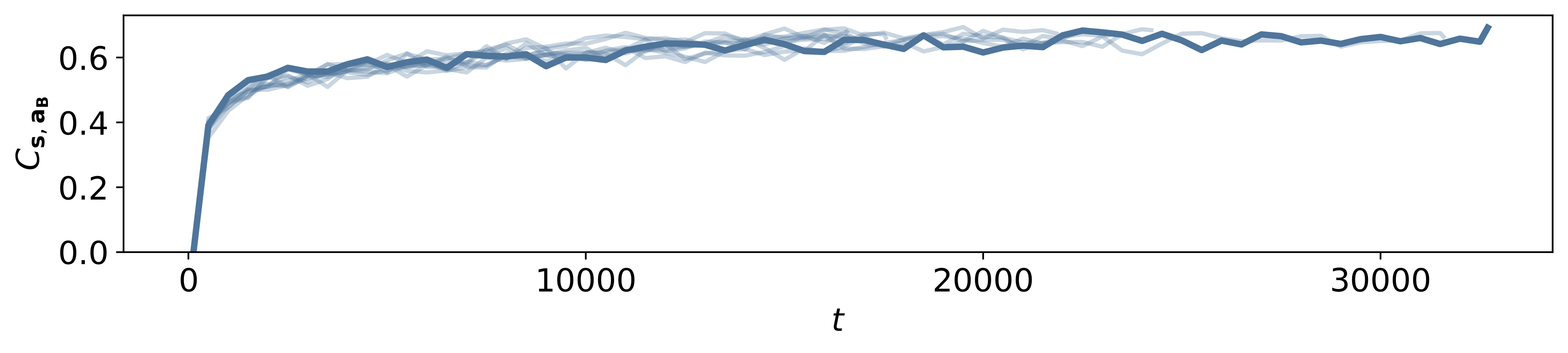}
    \\
     \includegraphics[width = 0.31\columnwidth]{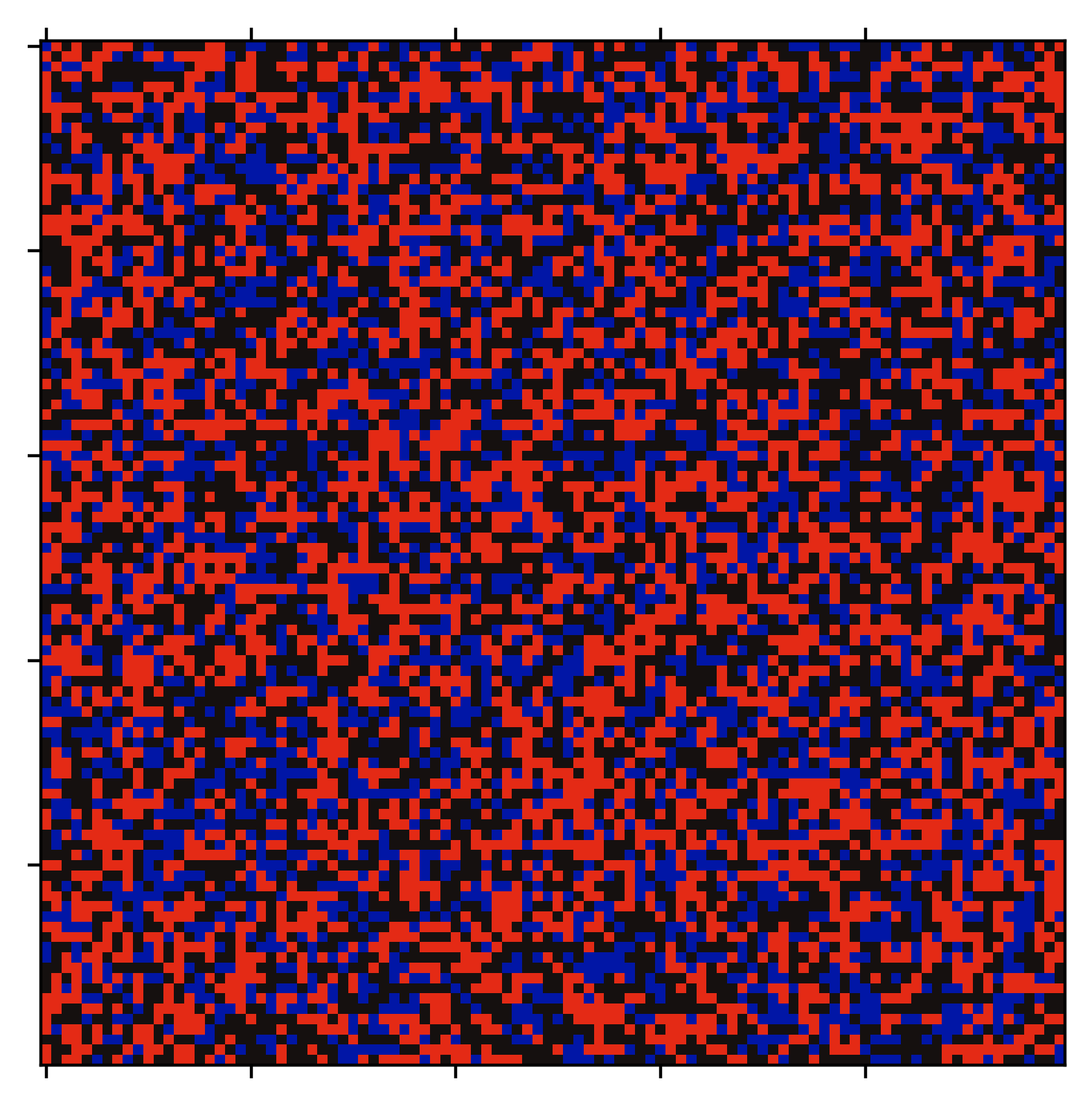}
     \includegraphics[width = 0.31\columnwidth]{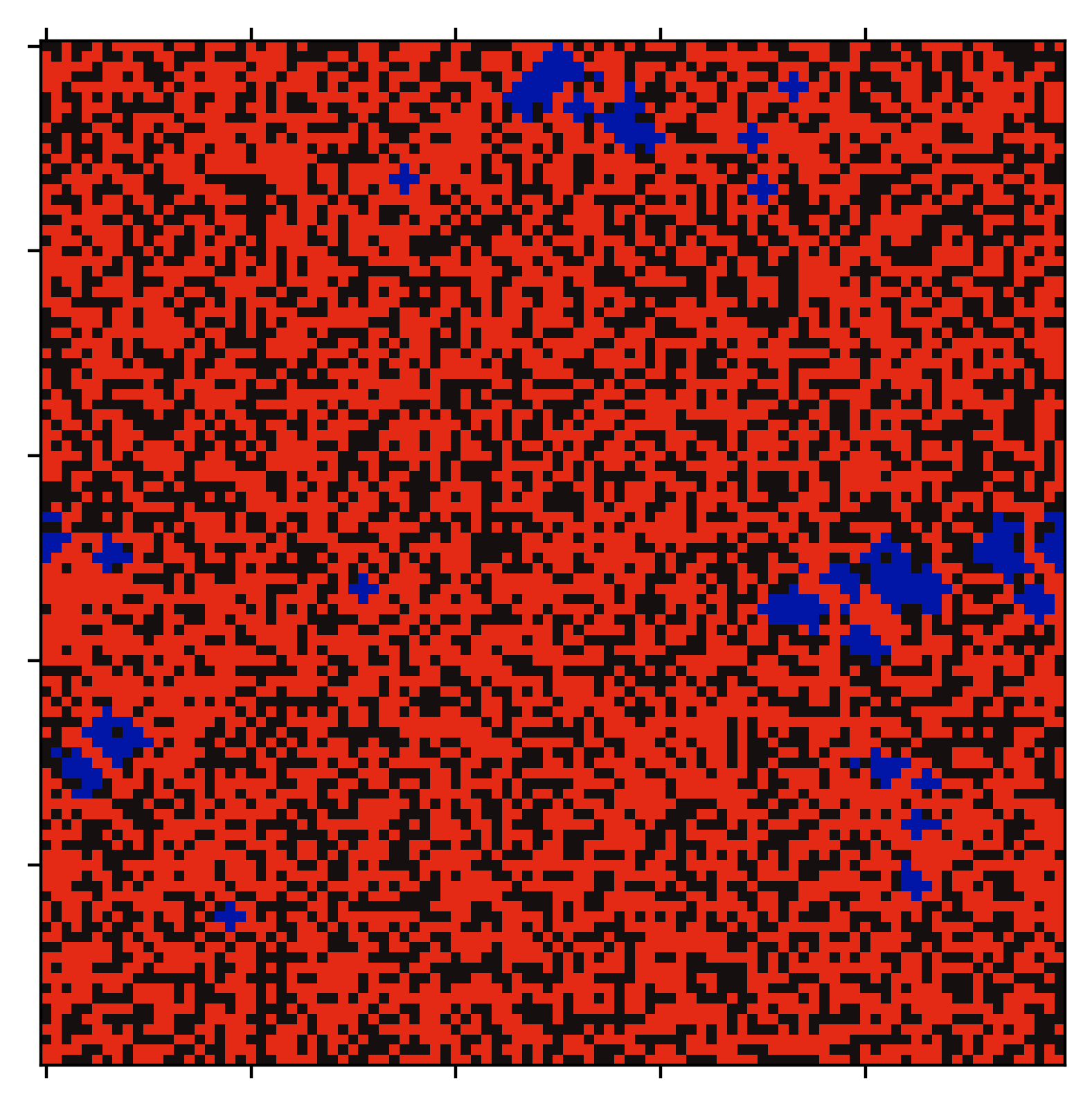}
     \includegraphics[width = 0.31\columnwidth]{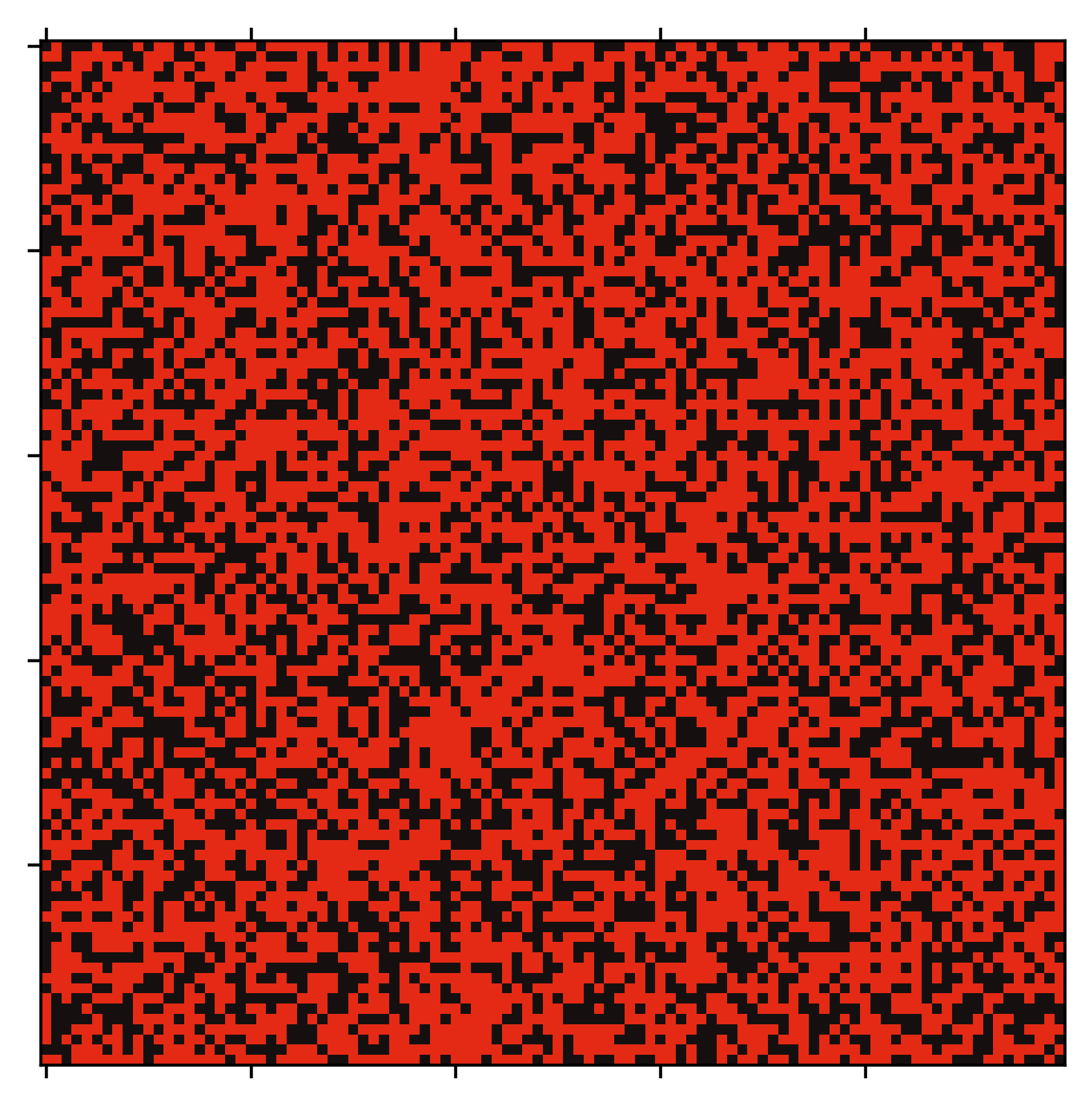} 
    \\
     \includegraphics[width = 0.31\columnwidth]{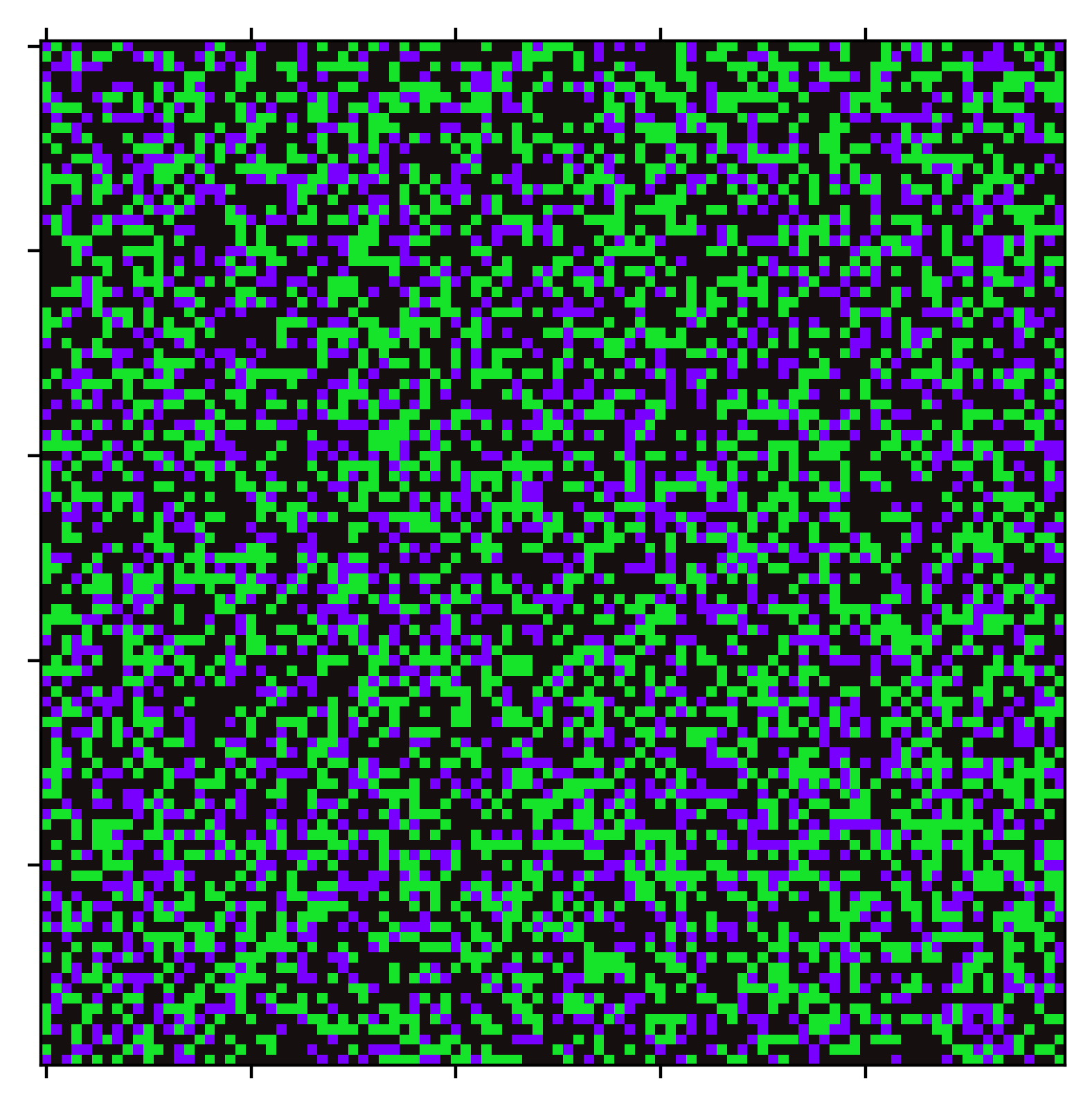}
     \includegraphics[width = 0.31\columnwidth]{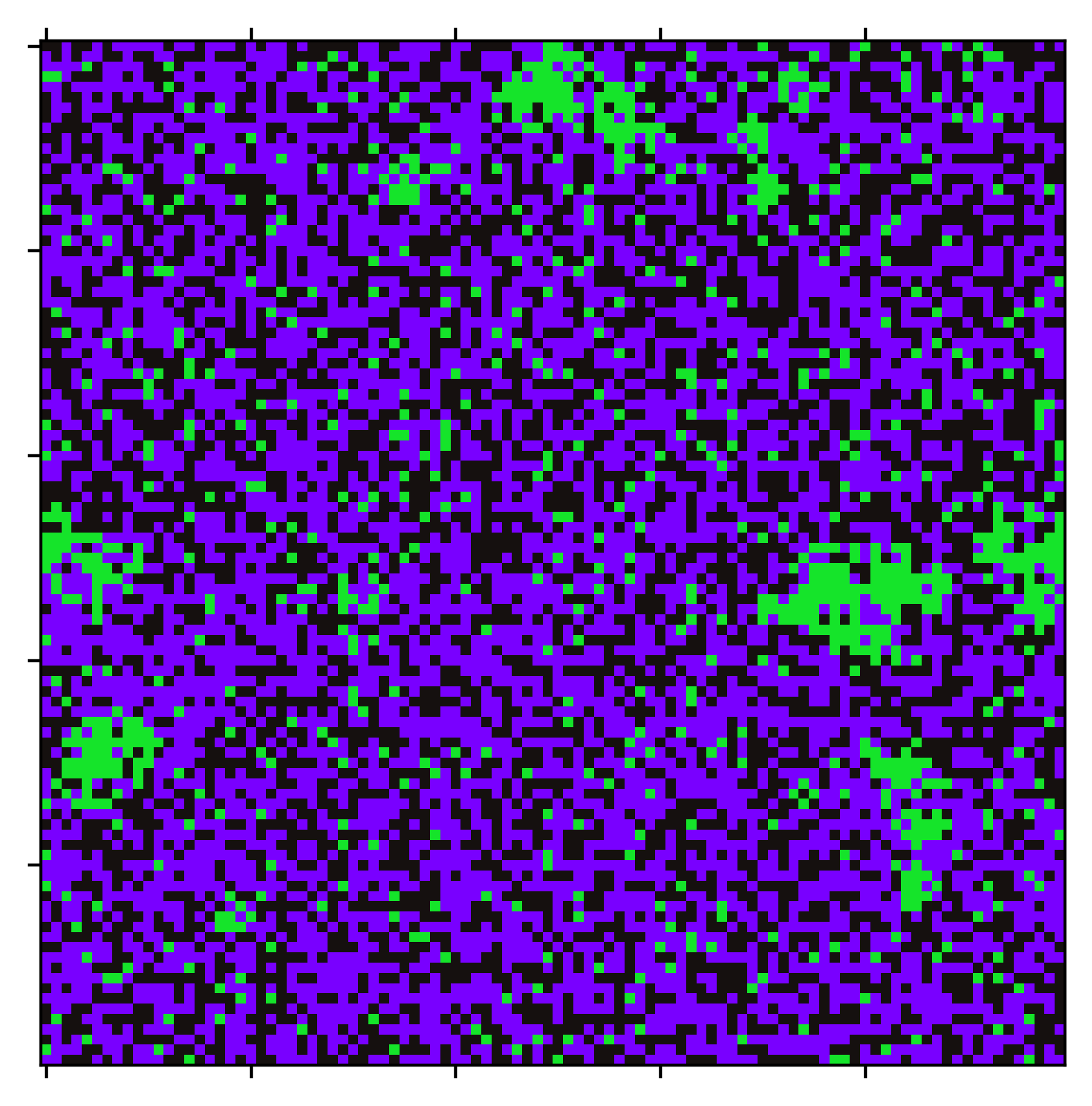}
     \includegraphics[width = 0.31\columnwidth]{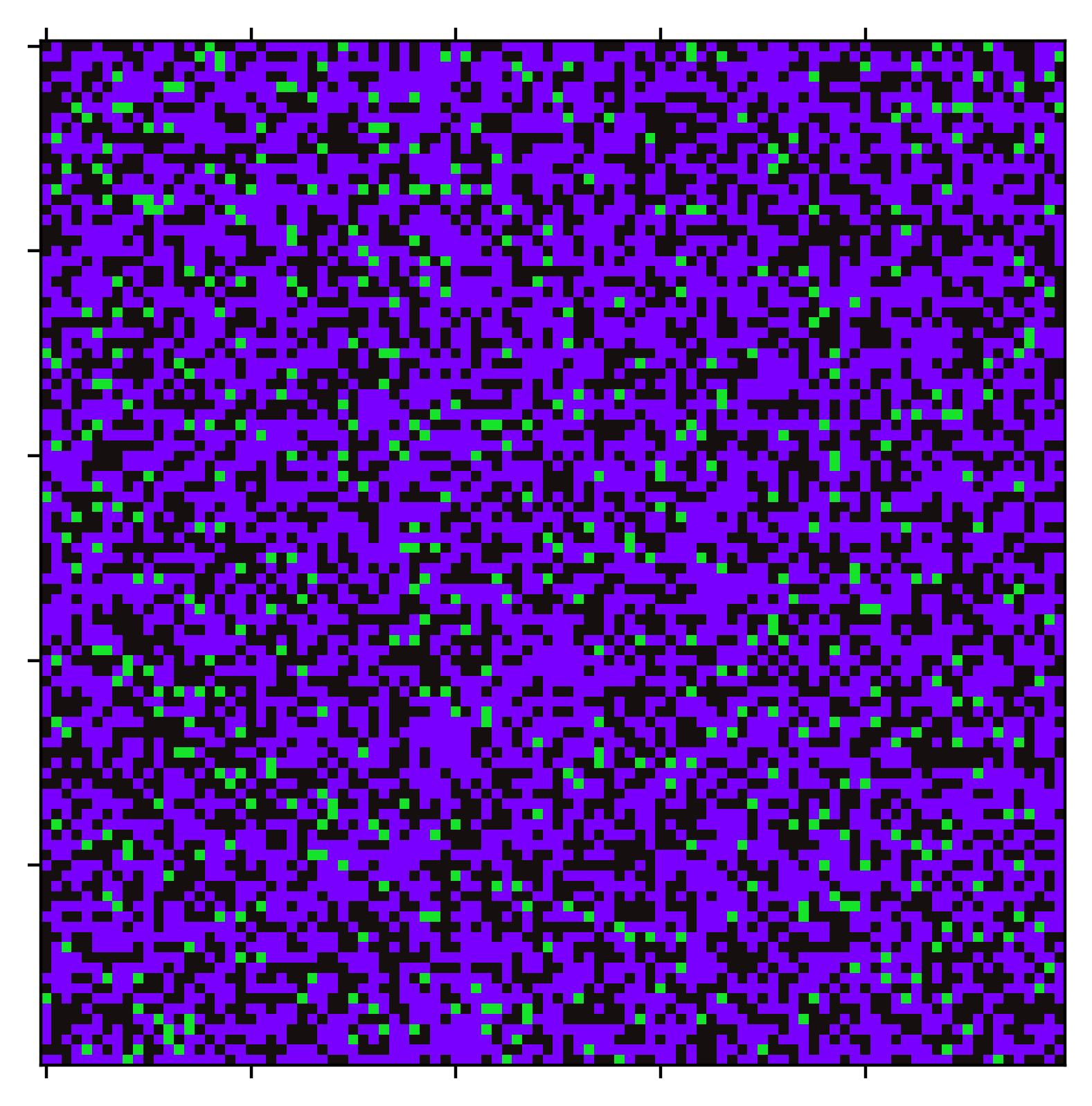} 
\caption{Typical snapshots showing both the state space $\mathbf{s}$ variables and the action space $\mathbf{a_B}$ variables, together with the correlation between the state space and action space shown in the first row. The showcased snapshots' sample is in bold color in the correlation plot, while additional samples are shown in the background. 
Time increases to the right, where the middle row is the state space, with cooperators in blue and defectors in red, and the last row is the action space, with \textit{copy-the-best} players in green while \textit{move} players are in purple, showing visually the correlation between those players and their respective roles as cooperators and defectors. Relevant parameters are $p_d=0.1$ and $\rho = 0.6$.
}
\label{fig:snaps-greatest=with-action-space}
\end{figure}

Furthermore, we show snapshots of the state and action variable values for the same configuration in Fig.~\ref{fig:snaps-greatest=with-action-space}, together with a time series correlation plot, in which the spatial correlation between the matrices that represent each space was measured using a simple Pearson correlation~\cite{Benesty2009}.
% colocar um highlight no regime que tem snapshots

A clear correspondence is evident in both the correlation data and the snapshots: players who choose action $B$ tend to form cooperative clusters, while those who choose action $M$ are typically defectors, showcasing the tendency of defectors to remain mobile and successfully invade cooperative groups. 
Therefore, if  $p_d$ is high, defectors are able to move around more freely and can easily attack cooperation clusters. At the moment of invasion, defectors achieve higher payoffs than cooperators by exploiting them, causing all players in the cluster to quickly switch to defection, breaking the spatial reciprocity mechanism. 
This analysis reveals a dynamic not previously reported in the literature and that is exclusive to using  the reinforcement learning framework.


%In the moment of invasion, the defectors' payoffs are higher then cooperators, as they are feeding off of them, so quickly all the players in the cluster turn to defection as well.
%This analysis produces a result that is not presented in literature and is exclusive to using the reinforcement learning framework, where we can separate the state and action spaces.
%It is also important to note this now, as we will discuss in the next section this situation when we add another type of action, which causes an increase in the endurance of cooperation through symbiosis.

In general, the result that low levels of mobility improve cooperation has been presented before~\cite{vainstein2007does}, with the important difference that those experiments were performed in synchronous fashion with fixed rules in the context of evolutionary game theory (i.e., without reinforcement learning).
In that case, %ese synchronous simulations with fixed update rules and thus not using reinforcement learning, 
 players performed a round of combats in parallel and then updated their states all at once by choosing the strategy of the best neighbour. 
 Movement %, in this case, 
  was introduced in an asynchronous fashion after the games were played, resulting in a curve that is qualitatively analogous to Fig.~\ref{fig:greatest-coop-curve}.

Our work makes entirely asynchronous, or, when looked at through the lenses of reinforcement learning, \textit{on-policy} updates. 
A common theme in both scenarios is the influence of the percolation threshold on cooperative agents. %—an effect that is not novel to this work. 
 This phenomenon has been extensively studied using both deterministic and stochastic update rules on diluted regular lattices~\cite{LI20211, wang2012if, PhysRevE.85.037101, PhysRevE.111.024123}, demonstrating how it can shape the outcome of social games.
 Apart of the novel results, we see that this already studied case with multi-agent reinforcement learning can help us produce results that serve as a benchmark, as we have done in this section by qualitatively reproducing findings from the literature, which in turn can enhance our understanding of the conditions under which convergence occurs in large-scale systems.

\subsubsection{{\label{sec:persist-or-choose-greatest}} Persist and copy-the-best}
\begin{figure*}[thbp]
\setkeys{Gin}{height=.75\textwidth}
\captionsetup[subfigure]{justification=centering}
    \centering
    \begin{subfigure}{0.499\textwidth}

     \includegraphics{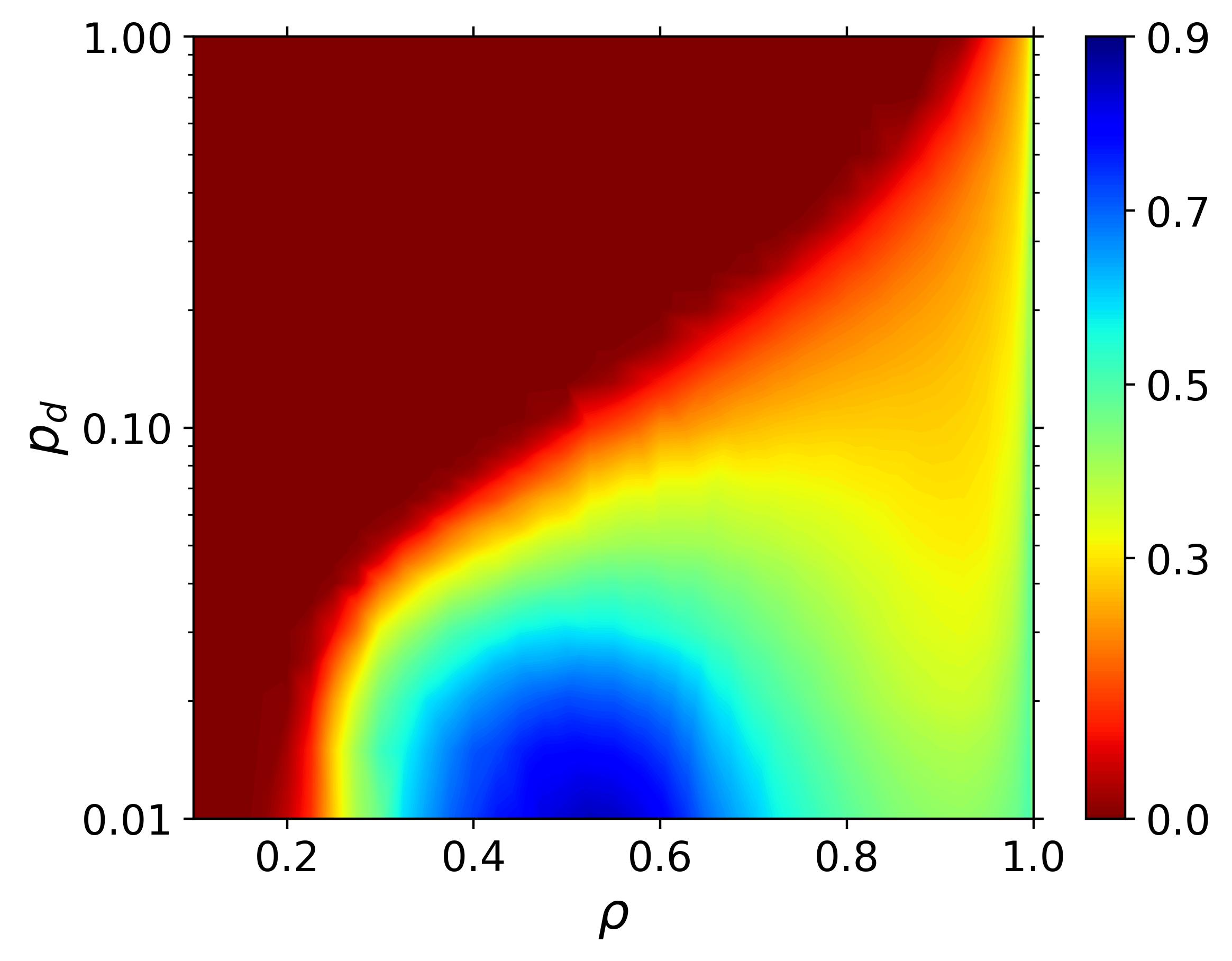} 
     \caption{}
     \label{fig:persist-or-choosing-the-greatest-curves}
    \end{subfigure}\hfill
    \begin{subfigure}{0.499\textwidth}
     \includegraphics{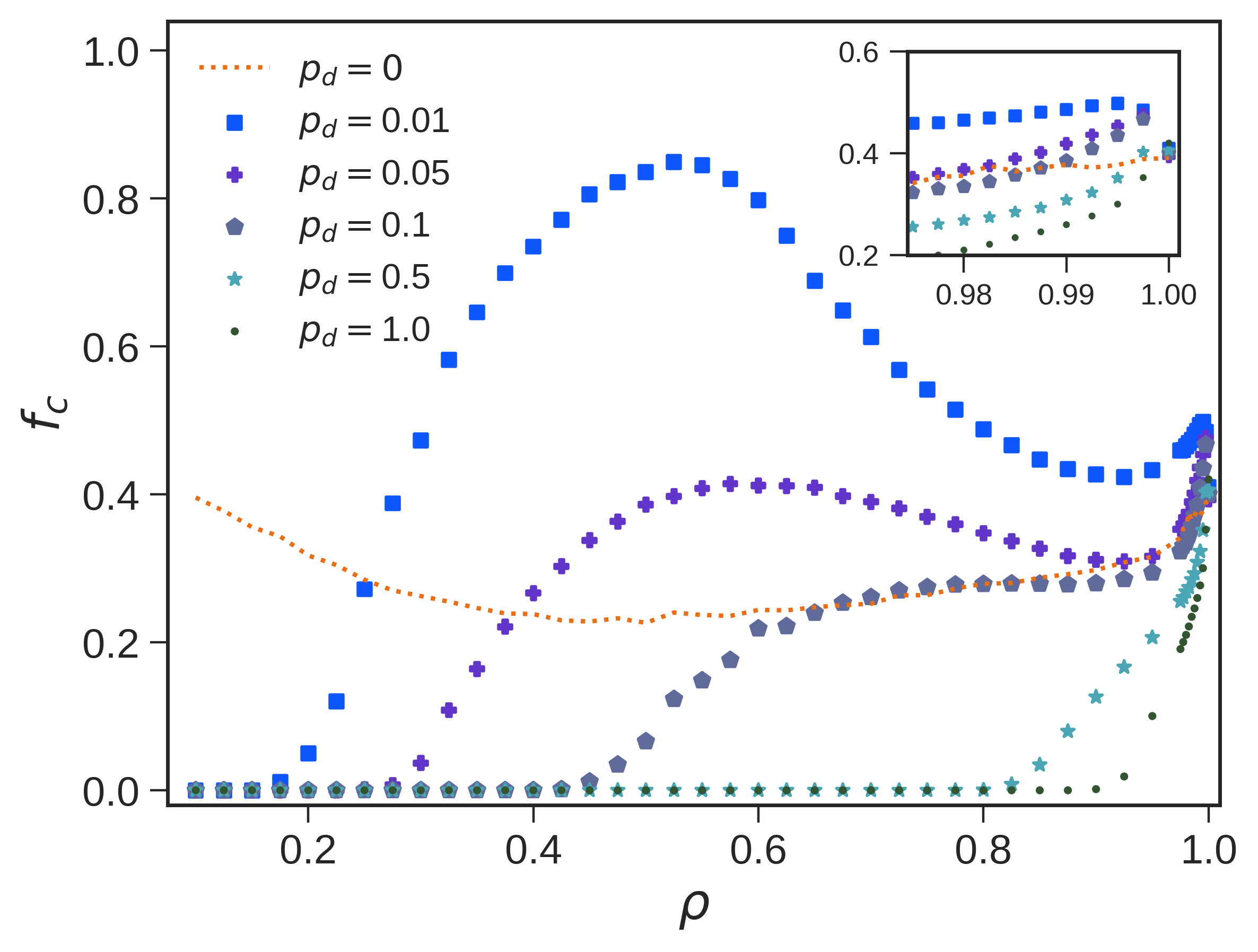}
     \caption{}
     \label{fig:stochastic-choosing-the-greatest-noise-heatmap}
    \end{subfigure}\hfill
\caption{Cooperation  \textbf{(a)} Heat map of the fraction of cooperators for different levels of mobility, varying together with the density of the lattice, where we see the same cooperative region shifted to the left. The axis $p_d$ is in logarithmic scale.  \textbf{(b)} Curves for specific values of the mobility rate, showcasing the differences and general increase in cooperation with the addition of the action $P$, or \textit{persist}. }
\label{fig:persist-or-choose-plots}
\end{figure*}
\begin{figure*}[thbp]
    \centering
    \includegraphics[width=.9\textwidth]{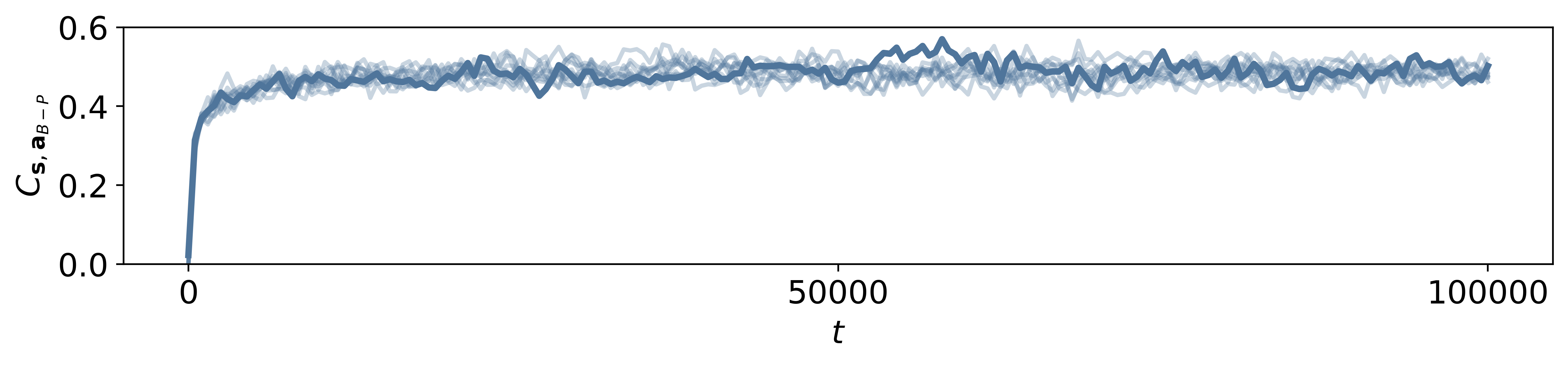}
    \\
     \includegraphics[width = .3\textwidth]{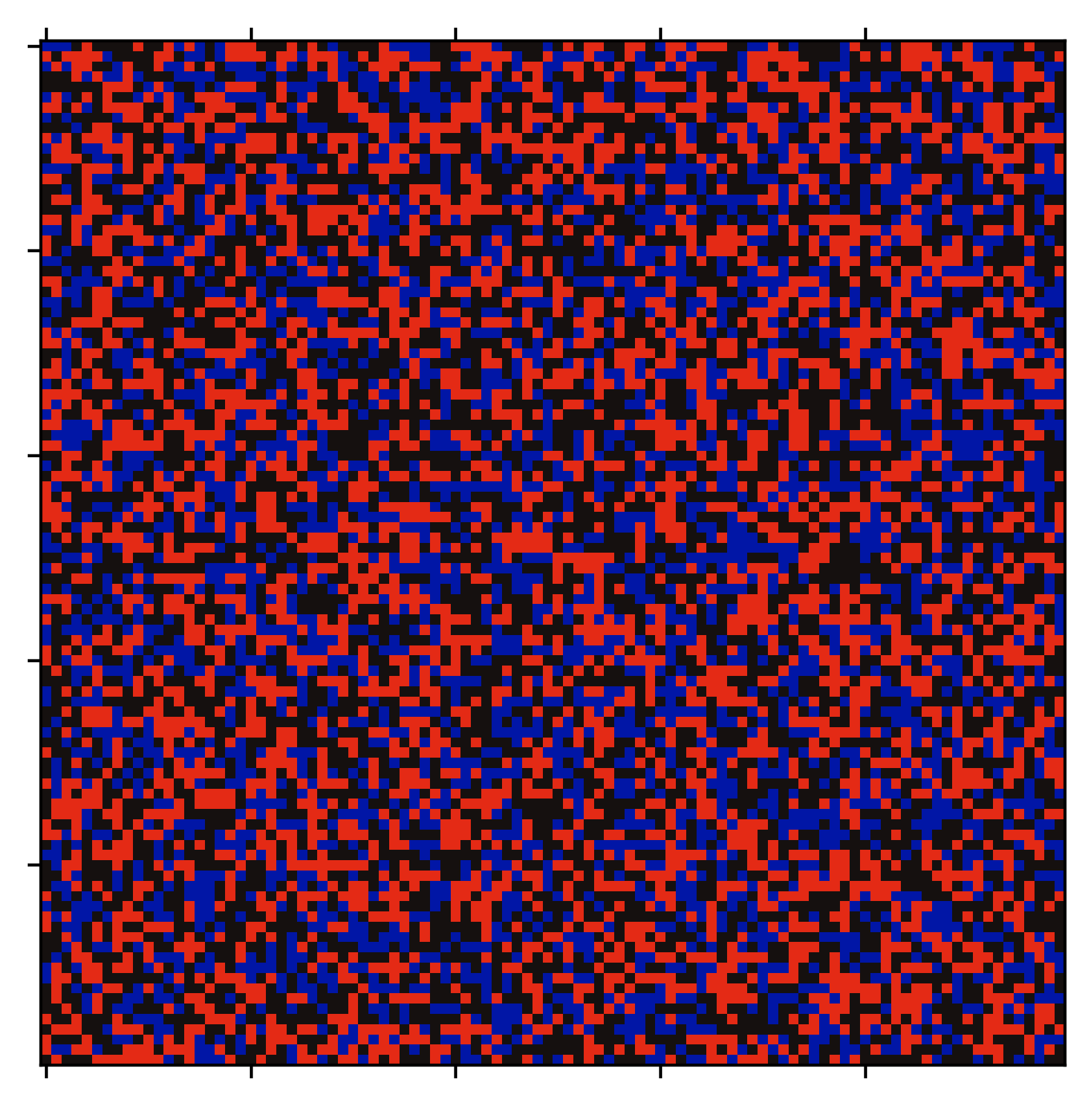} 
     \includegraphics[width = .3\textwidth]{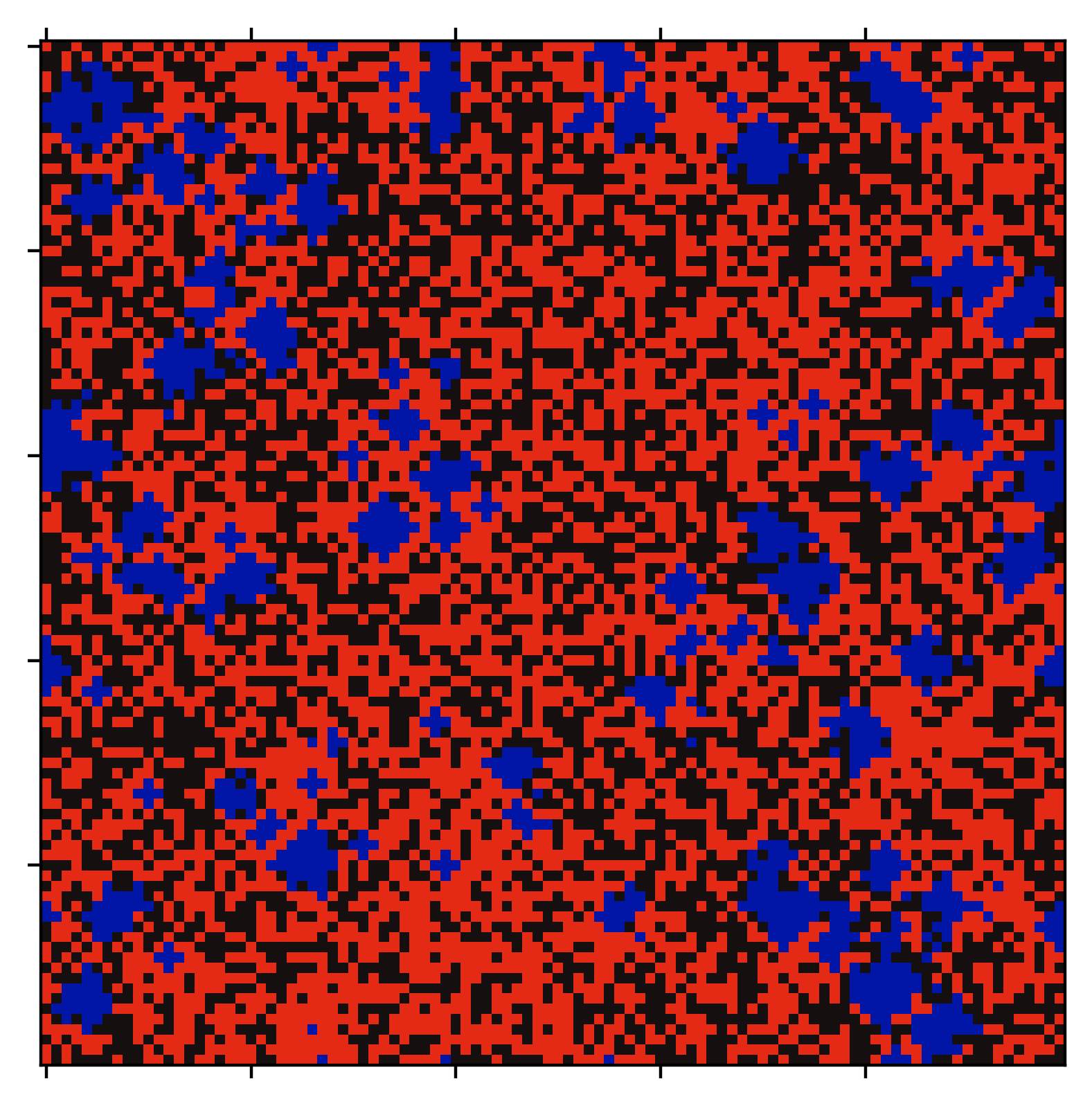}
     \includegraphics[width = .3\textwidth]{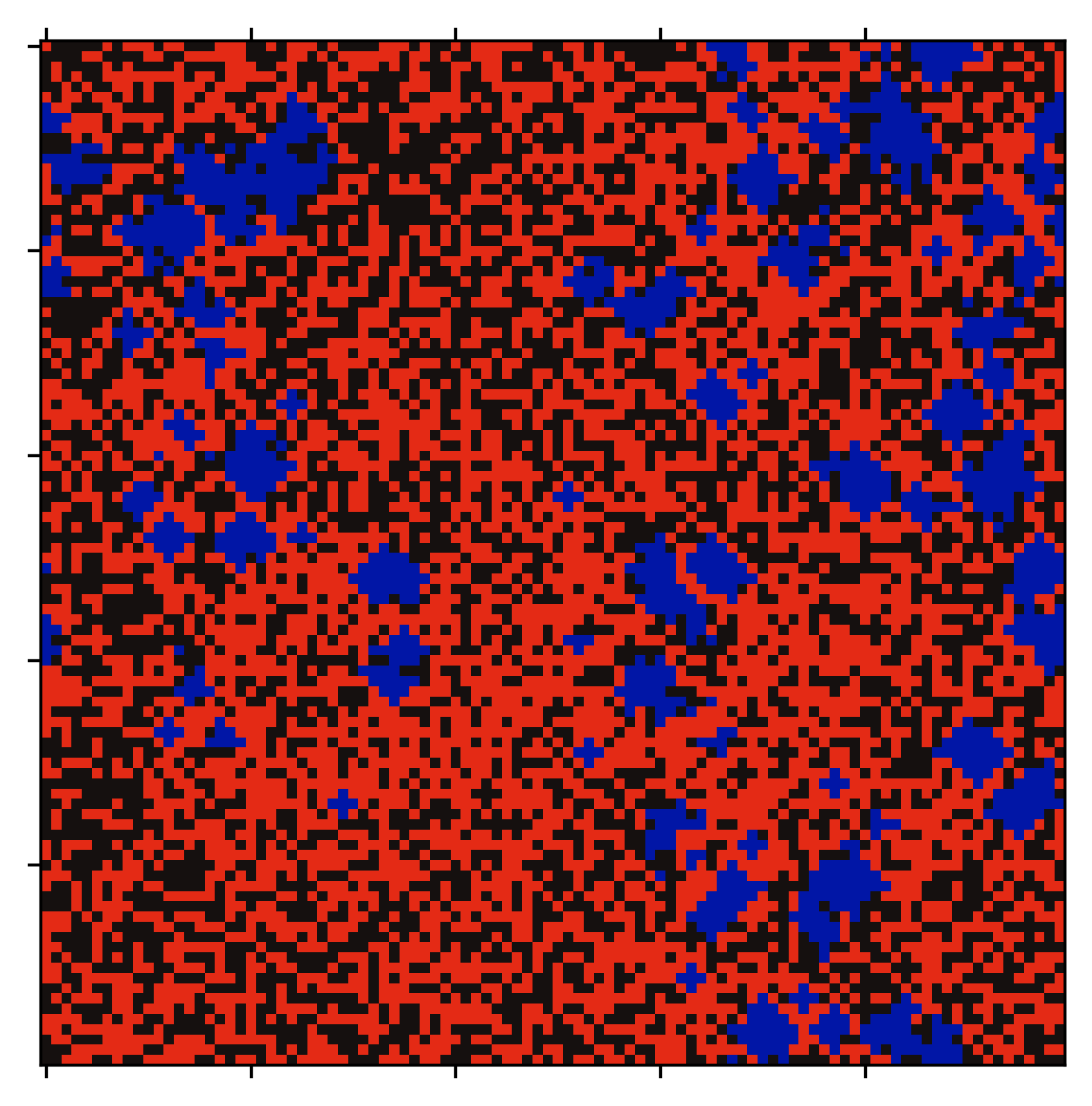} 
    \\     
     \includegraphics[width = .3\textwidth]{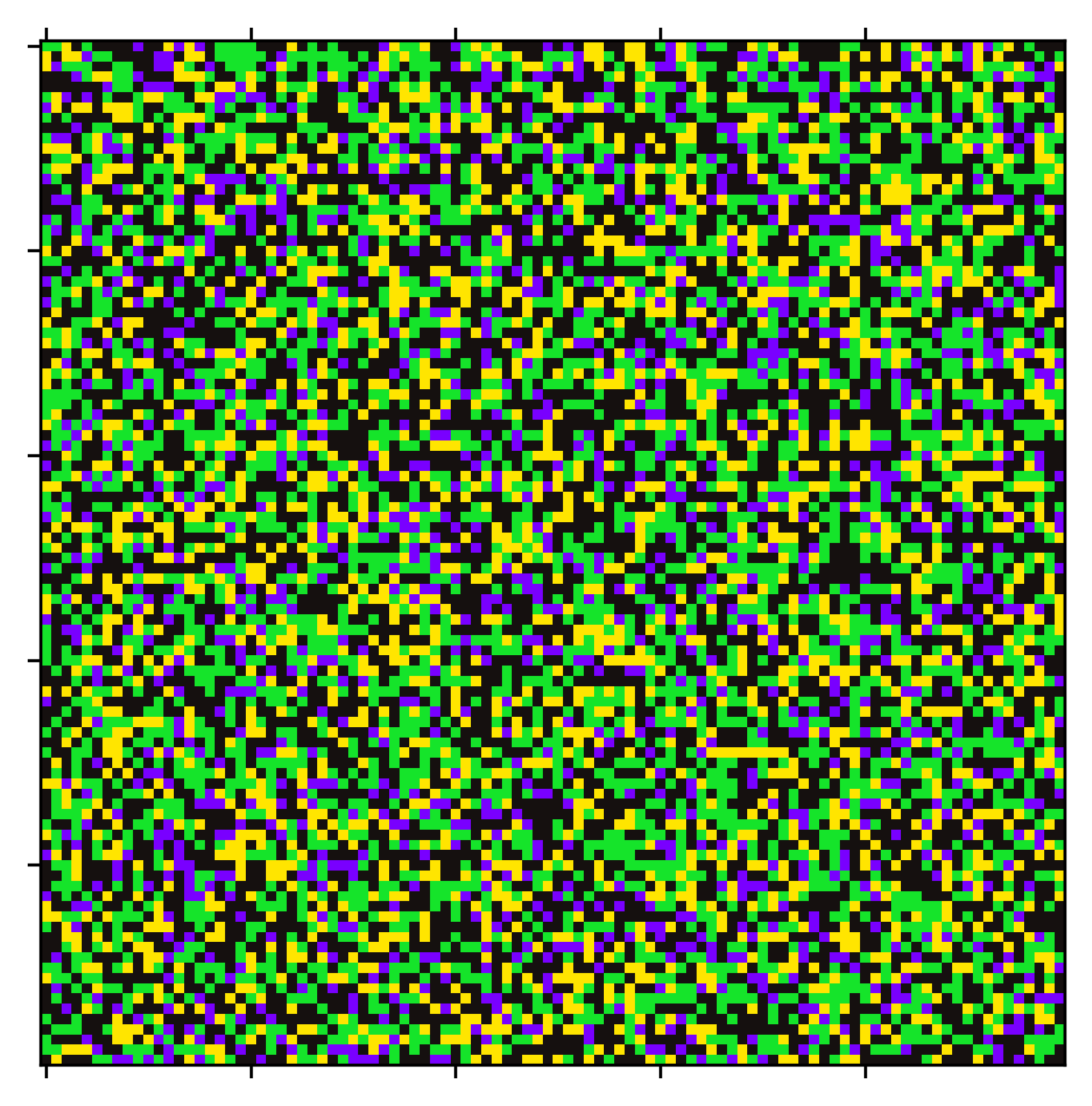} 
     \includegraphics[width = .3\textwidth]{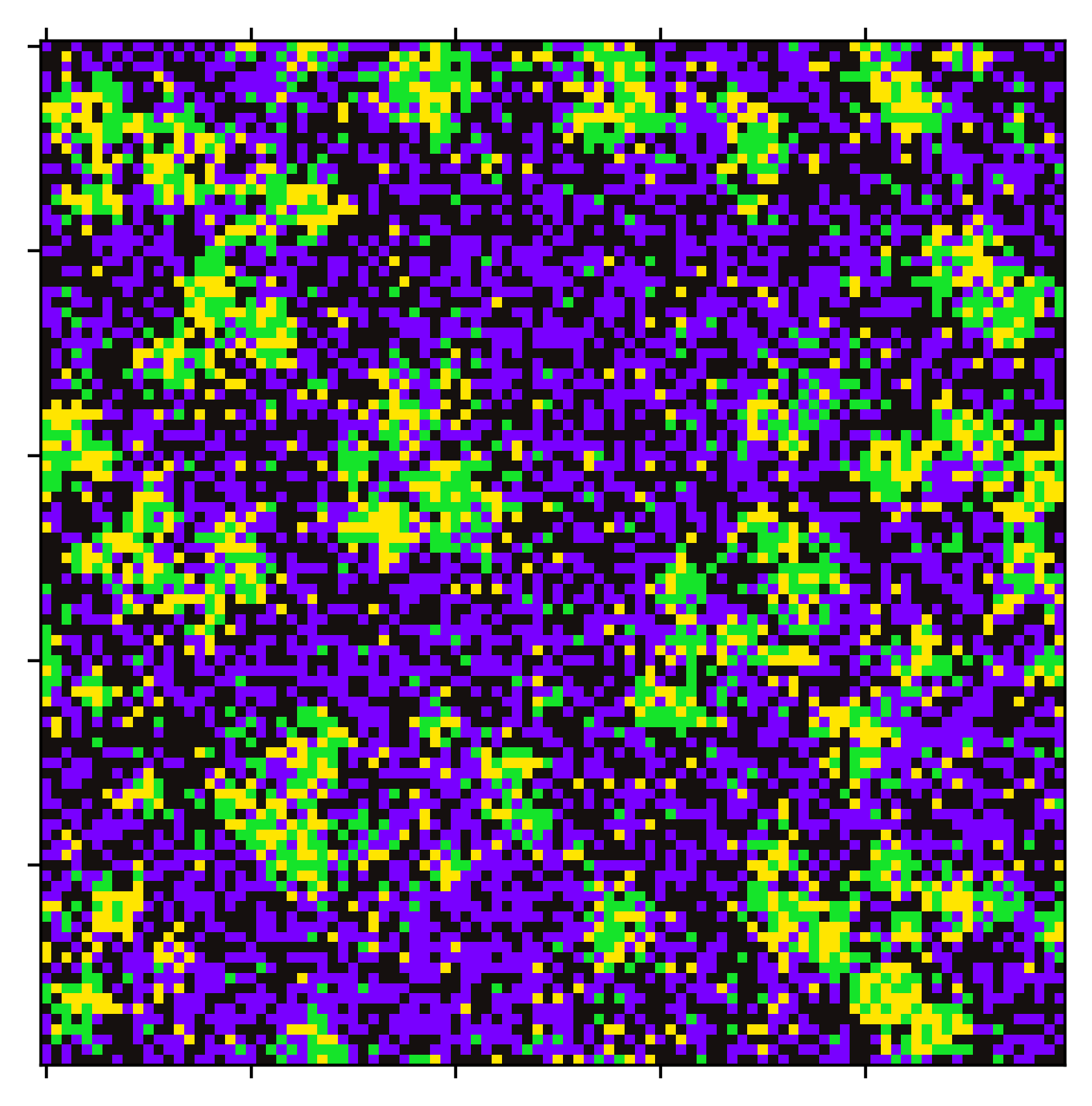}
     \includegraphics[width = .3\textwidth]{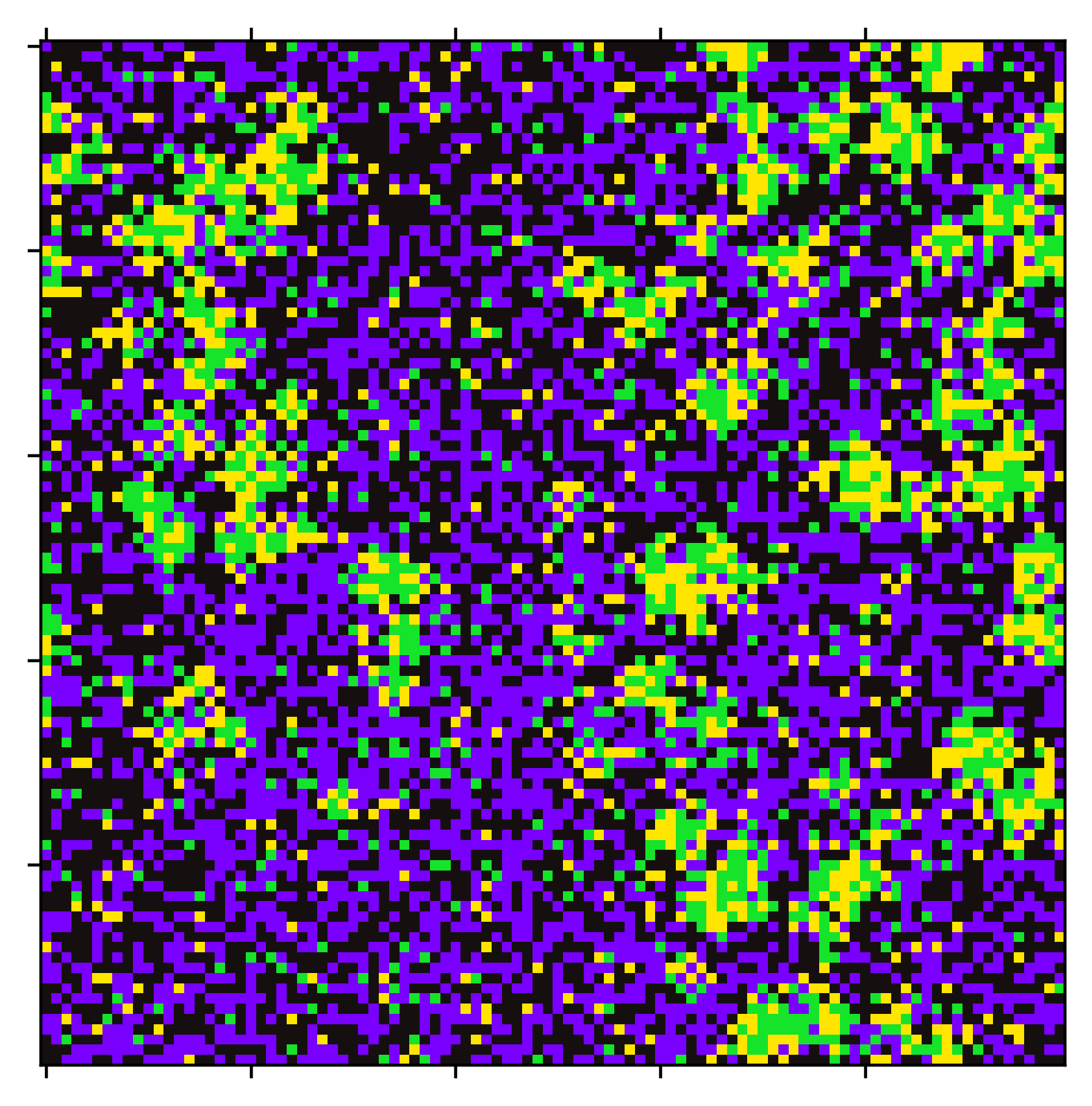} 
\caption{Typical snapshots from the simulation with the action set $\mathbf{a_{PB}}$, where the middle row shows the state space, with blue being $C$, red being $D$ and black being holes. The lower row shows holes in black as well, while the agents' actions space is represented in the color purple for $M$ or \textit{move}, in green for $B$ or \textit{copy-the-best} and yellow for $P$ or \textit{persist}. 
In the top row, we show the correlation between the state space and the action space for $10$ samples ran with the same parameters, showcasing the configuration from the snapshots in bold color.
The clear correspondence between the clusters can be seen, as well as the symbiotic mutualistic behaviour between $P$ and $B$ agents, which form the cooperation clusters, better resisting invasion from mobile agents $M$, which are mostly defectors.
}
\label{fig:snaps-greatest-persist}
\end{figure*}
%When looking at the snapshots in Fig.~\ref{fig:snaps-greatest=with-action-space}, we can also see why fast agents turn to defection, as cooperation clusters are largely formed by $B$ players who stay still and sustain the clusters by comparing themselves with other cooperating agents.
%As seen in the snapshots of the action space, defectors are also usually the ones that turn mobile and choose $M$, so, if $p_d$ is too high, defectors are able to move around more freely and can easily attack cooperation clusters, e.g., they can insert themselves into the neighbourhoods of clusters more easily and make the cooperators in it compare their payoffs with them.
%In the moment of invasion, the defectors' payoffs are higher then cooperators, as they are feeding off of them, so quickly all the players in the cluster turn to defection as well.
%This analysis produces a result that is not presented in literature and is exclusive to using the reinforcement learning framework, where we can separate the state and action spaces.
%It is also important to note this now, as we will discuss in the next section this situation when we add another type of action, which causes an increase in the endurance of cooperation through symbiosis.

Finally, we use an action set that involves three different actions %, as described in section \ref{sec:model}, giving us 
with our last Q-table:
\begin{equation}
\mathbf{Q_{PB}} = \ \begin{blockarray}{ c c c}
\begin{block}{[ c c c]}
  Q_{CP} & Q_{CB} & Q_{CM} \\
Q_{DP} & Q_{DB} & Q_{DM}  \\
\end{block}
\end{blockarray} \ .
\label{table:persist-or-choose}
\end{equation}
This results in a dynamic environment similar to the one described in the previous section, as evidenced by comparing the curves in Fig.~\ref{fig:persist-or-choosing-the-greatest-curves} with those in Fig.~\ref{fig:greatest-heatmap}.
As with the action set $\mathbf{a_B}$, there is a strong dependence on the mobility parameter $p_d$: slower agents tend to cooperate, while faster ones lean toward defection. However, in this case, the peaks are even more pronounced, and cooperation emerges in regions where it was previously absent.
As observed with the previous action set, peaks appear across all mobility levels near the network's percolation threshold.  %, $\rho \approx 0.593$.
Notably, this shift occurs only for mobile agents  as the static case continues to exhibit the same behaviour as before.
%With it, we have a dynamic environment that is %, in essence,
 %similar to the one from the previous section, as can be seen by comparing   %set of actions $\mathbf{a_B}$ used in (\ref{table:greatest}),  where payoffs are compared.
%\st{Distinctively, the addition of the \textit{persist} action induces changes in behaviour that are more prominent in certain limits, as we will discuss.}
%We first observe that 
%the curves in Fig.~\ref{fig:persist-or-choosing-the-greatest-curves} to the ones in  Fig \ref{fig:greatest-heatmap}. 
%As with the actions $\mathbf{a_B}$, we have a strong dependency on the mobility parameter $p_d$, with slower agents trending towards cooperation and fast ones choosing defection, but with even more pronounced peaks seen, and cooperation thriving in regions where before it did not.
%Similarly to the last action set, we have peaks for all mobility's approaching the percolation threshold of the network $\rho \approx 0.593$.
%We also see that this change happens only for mobile agents, as the static case remains with the same behaviour.
To understand this, we examine the region where cooperation was previously absent in the scenario involving only actions $B$ and $M$, but now shows substantial levels following the introduction of the $P$ action. 
More specifically, this corresponds to the low-mobility region just below the percolation threshold, where Fig.~\ref{fig:greatest-coop-plots} shows a phase of total defection for most mobility values, while Fig.~\ref{fig:persist-or-choose-plots} reveals the emergence of cooperators.
%To understand this, we look at the region where cooperation was extinguished in the case where we had only the $B$ and $M$, and now has considerable levels with the addition of $P$.  That is, more specifically, the low mobility region before the percolation threshold, where we see a total defection phase  for most mobility's in Fig.~\ref{fig:greatest-coop-plots} and the presence of cooperators in Fig.~\ref{fig:persist-or-choose-plots}. 

The snapshots in Fig.~\ref{fig:snaps-greatest-persist} reveal the system's behaviour in this limiting case, uncovering a striking result. We first observe that the clusters in the action space align with those in the state space with cooperative clusters being  composed of agents choosing actions $B$ and $P$. As also seen in Fig.~\ref{fig:snaps-greatest=with-action-space}, the defectors in this scenario are predominantly mobile agents selecting action $M$, which attempt to prey on the cooperative clusters. Interspersed among them are isolated agents committed to other available actions. This predatory behaviour by mobile agents might have led to the complete collapse of cooperation -- an outcome previously observed in this regime when $P$ was not present. However, the emergence of a symbiotic relationship between $B$ and $P$ agents enables the formation of more resilient cooperative clusters, which are better able to withstand defector invasions.
%In the snapshots present in Fig.~\ref{fig:snaps-greatest-persist} we investigate this limit's behaviour and notice a striking result.
%We begin by noting that the action space's clusters match the state space's ones, where the clusters of cooperators are thus formed by $B$ and $P$ and, as seen also in the snapshots in Fig.~\ref{fig:snaps-greatest=with-action-space}, the defectors in this case are mostly mobile agents choosing $M$ that try to prey on the cooperation cluster, with some loners in between stuck to the other available actions.
%This preying from mobile agents could have been completely successful with only type of agent, as seen before with the extinction of cooperators in the same limit, but the emergence of a symbiosis between these agents induces cooperators to form stronger clusters that can better resist the invasion from defectors.

This surprising mutualistic symbiotic behaviour~\cite{bronstein2015mutualism}
 can be attributed to the role of persistent agents (those choosing $P$) who act as a barrier for $B$ players. By surrounding the $B$ agents, they reduce their exposure to defectors when evaluating neighbouring strategies, thereby preserving cooperation.
In return, $P$ agents benefit from this arrangement by integrating into cooperative clusters, from which they derive higher payoffs than they would receive in isolation or from being in contact with defectors.
%\st{This dynamic is consistent with mutualistic symbiosis, as it aligns with the definition of mutualism: 'The ecological interaction between two or more species where each species has a net benefit.'} \cite{bronstein2015mutualism}.
%This surprising behaviour can be explained by the fact that the persistent agents, those who choose $P$, serve as a shield for the $B$ players that, when looking around for the best player, do not have as much contact with defectors as they did before.
%The $P$ agents also obtain an advantage from this relation, as they form cooperation clusters that they can feed off of and gain better payoffs than when being outside the cluster.
%This is coherent with it being a mutualistic symbiotic behaviour, as the definition of mutualism is given as ``\textit{The ecological interaction between two or more species where each species has a net benefit.}'' \cite{bronstein2015mutualism}.

This type of emergent symbiotic behaviour has been observed in evolutionary games, as already demonstrated in~\cite{FLORES2021110737}; however, the emergent dynamics described here are novel in two key aspects. 
First, it hints at the potential relationship between these types of players in general spatial prisoner's dilemma settings, with fixed or learned update rules. Second, it greatly showcases the \textit{population-policy equivalence}~\cite{bloembergen2015evolutionary} characteristic of the reinforcement learning algorithm in evolutionary dynamics, which lets us view the set of players that choose a given action as pertaining to a certain population.
That is, all the agents that choose $B$ can be viewed as a population $B$. 
Furthermore, when changing its strategy from, for example, $B$ to $P$, the agent changes from one population to the other.
This gives more meaning to the observed mutualism, as it can be seen as an emergent behaviour from the interaction between two highly dynamical populations.
% What mobility does, in this case, is pulling the low mobility cooperation peak to the left, closer to the percolation limit.
% This suggests that, slow agents that find one another decide more often to cooperate, even for lower densities, showing that mobility is much more effective for cooperation in a more stochastic scenario.
% The behaviour seen in detail in Fig.~\ref{fig:fermi-coop-curve-zoom} for mobility $p_d=0.05$ shows us that there is a transition between total defection and total cooperation in a region where the lattice is almost empty, which shows a threshold density where agents who find one another transition from a total defection phase to a total cooperation phase.
% It is also interesting to note that the lowest mobility simulated, $p_d=0.01$, attains almost always total cooperation, only decaying precisely after reaching the percolation limit.
%%%%%
%%%%%
%%%%%
\section{\label{sec:conclusion}Conclusions}
In this paper, we have extensively simulated different scenarios of dilution and mobility within a reinforcement learning algorithm that is simple, interpretable and light-weight, showing its  suitability for evolutionary dynamics in diffusive environments.
With it, we showed that dilution and mobility can greatly affect cooperation in spatial configurations, as already established in the literature~\cite{vainstein2001disordered, vainstein2007does}, but this time including the ability for each player to independently learn and take actions, which is inherently different from the case with fixed update rules.
We also showed many novel effects in the multi-agent reinforcement learning aspect, such as the effect of dilution in no-knowledge agents, the difference between knowledge being introduced in a deterministic versus a stochastic way, the effects of fast and slow movement and the striking emergent mutualistic behaviour in persist-compare clusters.
Many open questions are left, of course, such as the role of asymmetry in interactions, which appears in our work when holes are present and not all agents have the same number of neighbours, as well as the effects of different types of movements, such as Lévy flights~\cite{viswanathan2000levy}, to name a few.

It is important to note, also, that the change in actions that produces a new set is arbitrary.
Actions to move in different ways, where the agent can choose to move preferably in  certain directions or to move in a non-diffusive manner in general,
%acho que no tcc da fernanda tinha umas refs sobre isso
are other examples of applications that the reinforcement learning framework can greatly help. % with greatly.
We also highlight the most significant distinction between the classical spatial games approach and the reinforcement learning framework in the same setting, which is the introduction of  choice. 
In the latter, players are given the agency to choose among actions and learn from the outcomes, introducing  a certain level of adaptation that is intrinsic to the player's perspective.
This is clear  in the form of a result that appeared in literature but was not discussed \cite{wang2022levy}, which is the non-vanishing of cooperation for higher temptation in the no-knowledge case.
In fixed update rules settings such as using choose the best or the Fermi rule, when players do not learn, cooperation quickly goes to zero \cite{nowak1992evolutionary, perc2008social}.

% essa parte foi adicionada, estava no texto
Furthermore, our work highlights another important aspect: the convergence challenges of multi-agent reinforcement learning algorithms. These problems often arise due to the non-stationary nature of the environment, which invalidates the convergence guarantees typically associated with single-agent reinforcement learning~\cite{busoniu2008comprehensive,canese2021multi}, such as the loss of the Markov property from the perspective of each agent. 
Various strategies have been proposed to address this issue~\cite{papoudakis2019dealing}, particularly involving alternative algorithms that are not inherently independent and are generally applied in settings with a small number of agents.
However, performing simulations with a large number of agents, such as in our study, poses significant challenges in terms of convergence, which our work deals with by falling back into known~\cite{vainstein2007does} and now benchmarked results.

Broadening the field of open questions and applications, we note that although Q-learning is exact and ideal for these low dimensional state and action spaces we used, this might not always be the case, and newer algorithms can be used, always leveraging computational power, interpretability and suitability to the models studied.
%As argued before, we can point out the use of these known results, such as the one reproduced qualitatively in this paper on the role of mobility in cooperation, can help benchmark how multi-agent reinforcement learning can converge in large systems such as these, as the independent learning algorithm utilized here shows.
In the same vein of benchmarking, many themes in single or multi-agent reinforcement learning theory can be studied and exemplified by simple and already studied concepts in game theory, as shown by~\cite{lanctot2017unified, rajeswaran2020game, yang2020overview, de2024game}, and we believe that the state-action modelling done in our work contributes to different frameworks, involving mobility or not, and that the reinforcement learning community may use such simulations to advance the studies in the algorithms themselves, understanding, for example, how a large number of learning agents learn to coordinate, which are the conditions for them to do so and how cooperation plays a role in it.

\section*{Acknowledgements}
G.C.M. thanks the Brazilian funding agency CAPES for the M. Sc. scholarship.
H.C.M.F. and M.H.V. acknowledge the financial support from the National Council for Scientific and Technological Development – CNPq (proc. 402487/2023-0). The simulations were conducted using the \href{https://pnipe.mcti.gov.br/laboratory/19775}{VD Lab} %,  \url{https://pnipe.mcti.gov.br/laboratory/19775},  
 cluster infrastructure at IF-UFRGS. 

\section*{Declaration of generative AI and AI-assisted technologies in the writing process}
During the preparation of this work the author(s) used ChatGPT-o1 in order to detect and fix grammar mistakes. After using this tool/service, the author(s) reviewed and edited the content as needed and take(s) full responsibility for the content of the published article.
\newpage
\bibliographystyle{apsrev4-2}
\bibliography{bib}

\end{document}